\documentclass[10pt,twocolumn,letterpaper]{article}

\usepackage[pagenumbers]{cvpr} 

\usepackage{url}
\usepackage[utf8]{inputenc} 
\usepackage[T1]{fontenc}    
\usepackage{url}            
\usepackage{booktabs}       
\usepackage{amsfonts}       
\usepackage{nicefrac}       
\usepackage{microtype}      
\usepackage{xcolor}    
\usepackage[most]{tcolorbox}
\PassOptionsToPackage{options}{natbib}
\usepackage{graphicx}
\usepackage{makecell}
\usepackage{booktabs}
\usepackage{multirow}
\usepackage{array}
\usepackage{wrapfig}
\usepackage{adjustbox}
\usepackage{subcaption}
\usepackage{caption}
\usepackage{colortbl}
 \usepackage{pifont}
\usepackage[ruled,vlined,linesnumberedhidden]{algorithm2e}
\usepackage{enumitem}
\usepackage{pifont}
\newcommand{\xmark}{\ding{55}} 
\newcommand{\cmark}{\ding{51}} 
\usepackage{amsmath}
\usepackage{textcomp}

\newcommand{\datasetname}[1]{MiLDEBench}
\newcommand{\methodname}[1]{MiLDEAgent}
\newcommand{\evalmetrics}[1]{MiLDEEval}
\newcommand{\evalscore}[1]{MiLDEScore}
\newcommand{\contenteditscore}[1]{$Score_{CE}$}
\newcommand{\layouteditscore}[1]{$Score_{LE}$}

\newlist{steplist}{enumerate}{1}
\setlist[steplist,1]{
  label=\arabic*.,      
  left=1.5em,           
  labelsep=0.5em,       
  itemsep=0pt,          
  topsep=2pt,           
  parsep=0pt,           
  align=left,           
  nosep                 
}

\definecolor{backred}{RGB}{255, 190, 190}
\definecolor{backblue}{RGB}{210, 230, 250}

\definecolor{darkblue}{rgb}{0.0, 0.0, 0.55}
 
\definecolor{cvprblue}{rgb}{0.21,0.49,0.74}
\usepackage[pagebackref,breaklinks,colorlinks,allcolors=cvprblue]{hyperref}

\def\paperID{14949} 
\def\confName{CVPR}
\def\confYear{2026}

\title{MiLDEdit: Reasoning-Based Multi-Layer Design Document Editing}

\author{
Zihao Lin\textsuperscript{1}\footnotemark \quad
Wanrong Zhu\textsuperscript{2} \quad
Jiuxiang Gu\textsuperscript{2} \quad
Jihyung Kil\textsuperscript{2} \quad
Christopher Tensmeyer\textsuperscript{2} \\
Lin Zhang\textsuperscript{3} \quad
Shilong Liu\textsuperscript{4} \quad
Ruiyi Zhang\textsuperscript{2} \quad
Lifu Huang\textsuperscript{1} \quad
Vlad I. Morariu\textsuperscript{2} \quad
Tong Sun\textsuperscript{2} \\[0.5em]
\textsuperscript{1}University of California, Davis \quad \textsuperscript{2}Adobe \quad
\textsuperscript{3}UW-Madison \quad
\textsuperscript{4}Princeton University 
}

\begin{document}
\renewcommand{\thefootnote}{\fnsymbol{footnote}}
\twocolumn[{
    \renewcommand\twocolumn[1][]{#1}
    \maketitle
    
    \begin{center}
        \parbox{\textwidth}{
            \includegraphics[width=\linewidth]{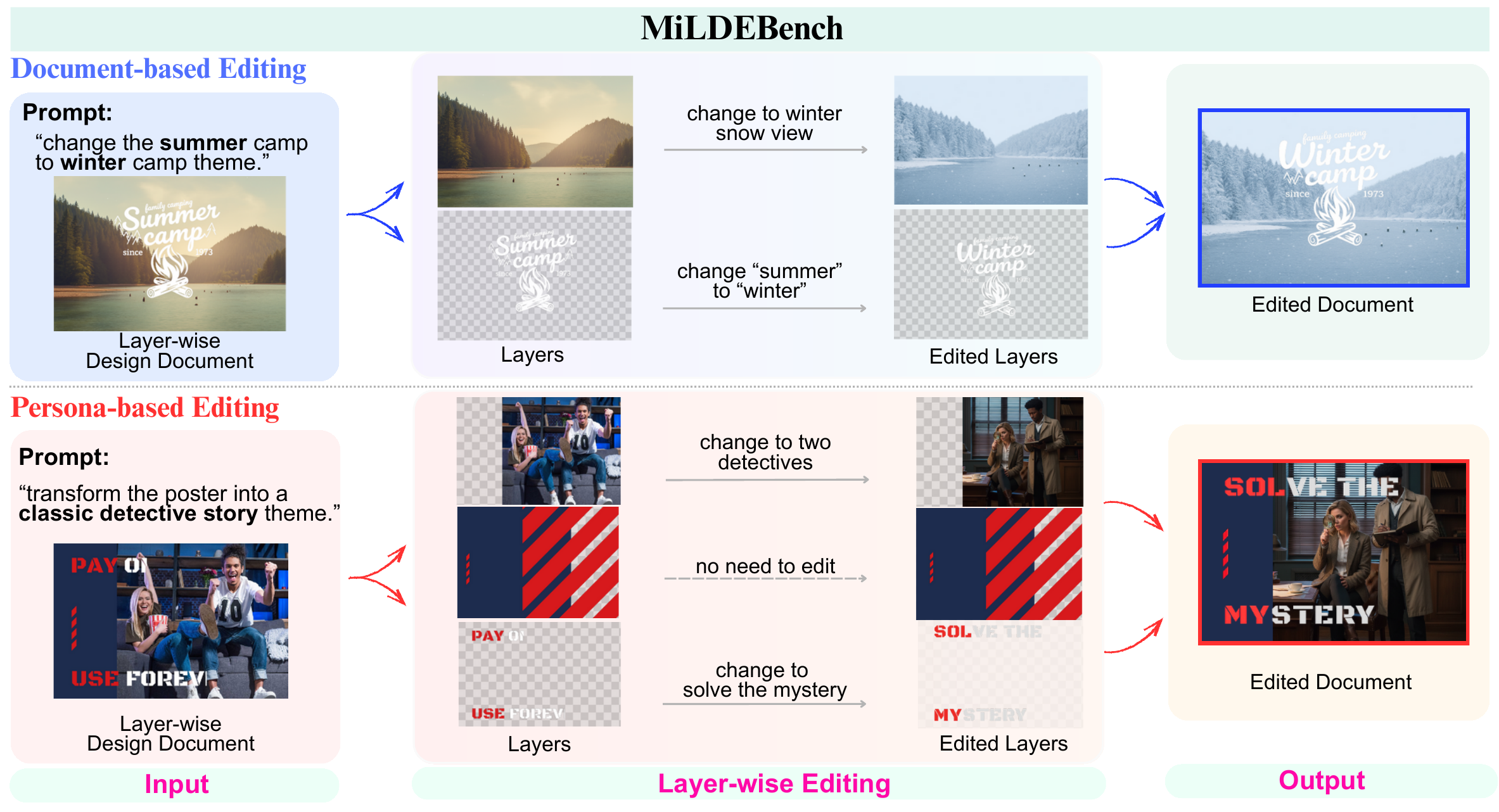} 
            \captionof{figure}{Examples of \datasetname~. Our benchmark is the first targeting to transparent-background, multi-layer design document editing.}
            \label{fig:multi-layer-editing}
        }
    \end{center}
}]
\footnotetext{* Corresponding author. Email: qzlin@ucdavis.edu. Work done during an internship at Adobe Research.}
\begin{abstract}
Real-world \textit{design documents} (e.g., posters) are inherently multi-layered, combining decoration, text, and images. Editing them from natural-language instructions requires fine-grained, layer-aware reasoning to identify relevant layers and coordinate modifications. Prior work largely overlooks \textit{multi-layer design document editing}, focusing instead on single-layer image editing or multi-layer generation, which assume a flat canvas and lack the reasoning needed to determine \textit{what} and \textit{where} to modify.
To address this gap, we introduce the \underline{M}ult\underline{i}-\underline{L}ayer \underline{D}ocument \underline{E}diting \underline{Agent} (\textbf{\methodname}), a reasoning-based framework that combines an RL-trained multimodal reasoner for layer-wise understanding with an image editor for targeted modifications. 
To systematically benchmark this setting, we introduce the \underline{M}ult\underline{i}-\underline{L}ayer \underline{D}ocument \underline{E}diting \underline{Bench}mark (\textbf{\datasetname}), a human-in-the-loop corpus of over 20K design documents paired with diverse editing instructions. The benchmark is complemented by a task-specific evaluation protocol, \textbf{\evalmetrics}, which spans four dimensions including instruction following, layout consistency, aesthetics, and text rendering.
Extensive experiments on 14 open-source and 2 closed-source models reveal that existing approaches fail to generalize: open-source models often cannot complete multi-layer document editing tasks, while closed-source models suffer from format violations. In contrast, \methodname~ achieves strong layer-aware reasoning and precise editing, significantly outperforming all open-source baselines and attaining performance comparable to closed-source models, thereby establishing the first strong baseline for multi-layer document editing. \vspace{2mm}
\end{abstract}

\section{Introduction}
\label{sec:intro}

While recent breakthroughs in image generation have transformed creative workflows, editing real-world design documents such as posters, flyers, and slides still remains an open challenge. 
Unlike natural images, these design documents are intrinsically multi-layered, combining backgrounds, graphics, text, and foreground imagery in a carefully structured hierarchy. Effective editing requires reasoning about which layers are relevant to user intent, how their relationships constrain possible modifications, and where changes can be applied without disrupting layout or occluding critical content. 
Existing reasoning-based editing methods \citep{jiang2025t2i, chen2025r2i, zhang2025reasongen} are built for flat, single-layer canvases and fail to capture this complexity. Besides, despite some works focuing on design document generation \citep{huang2024layerdiff, pu2025art, chen2025prismlayers}, layer-aware document editing remains unexplored, leaving a critical gap in vision-language reasoning and multimodal editing.

To fill this gap, we propose the first benchmark for reasoning-based \emph{multi-layer} document editing, \underline{M}ult\underline{i}-\underline{L}ayer \underline{D}ocument \underline{E}diting \underline{Bench}mark (\textbf{\datasetname}). \datasetname~ systematically focuses on \emph{content editing}, which entails semantically coherent modifications while maintaining the visual and structural integrity of the document. Building upon 20K transparent-background templates from the public Crello dataset \citep{yamaguchi2021canvasvae}, we synthesize 50K natural-language editing instructions and 87K layer-aligned edit steps via a hybrid generation pipeline that integrates open-source multimodal LLMs with human-in-the-loop verification.
To approximate real-world application scenarios where users come from diverse backgrounds, we design persona-conditioned and document-conditioned prompts that capture heterogeneous editing intents, ensuring that the dataset reflects a broad spectrum of user needs
(\eg, converting a Christmas card into a Halloween card).

To evaluate this new setting, we introduce \textbf{\evalmetrics}, a task-specific evaluation protocol that encompasses four core dimensions: \emph{instruction following}, \emph{layout consistency}, \emph{aesthetics}, and \emph{text rendering}. Together, these dimensions establish a standardized and comprehensive testbed for reasoning-intensive, layer-aware image editing, which closely mirrors real-world multi-layer design document editing scenarios. Furthermore, to better align the evaluation with human perceptual judgment, we aggregate the four criteria into a unified metric, termed \textbf{\evalscore}. This composite score provides a more holistic assessment of editing quality and demonstrates stronger correlation with human preference compared to previous evaluation protocols or any individual criterion.
We evaluate 14 open-source and 2 closed-source image-editing models on \datasetname~. Since most existing models can only produce a single edited output, we simplify our benchmark to a single-round image-editing task rather than a multi-layer editing scenario. Concretely, each model receives one design document and one editing instruction, and is required to generate a single edited poster (without access to layer-level structural information). Even under this simplified setting, open-source models demonstrate limited instruction-following ability, frequently returning partially edited outputs. In contrast, closed-source models achieve higher semantic alignment and visual quality but sometimes compromise layout or format consistency. Incorporating explicit reasoning yields only modest improvements, indicating that existing reasoning modules are largely text-centric and do not fully leverage the multi-layer document structure. These findings suggest that multi-layer design document editing poses challenges beyond the scope of current image-editing paradigms and motivate the need for a reasoning-based, layer-aware approach.

To address these limitations and enable faithful multi-layer editing, we introduce \textbf{\methodname}, a reasoning-based, layer-aware editing agent. \methodname~ integrates (i) an RL-trained multimodal reasoner optimized with a novel reward function for layer identification and layer-conditioned prompt synthesis, and (ii) a modular image editor for targeted, layer-specific modifications. Experimental results demonstrate that explicit layer-aware reasoning is crucial for accurate and controllable document-level editing. Our method surpasses all open-source baselines by around 82.78\% in \evalscore~, achieves comparable performance compared to closed-source models, and further outperforms them in layout consistency. Notably, \methodname~ achieves the best balance between \emph{instruction adherence} and \emph{layout consistency}, underscoring the efficacy of reasoning-based multi-layer editing.

We summarize our main contribution as follows:

\begin{itemize}
\item \textbf{Task and Benchmark.} We formalize the problem of \emph{multi-layer design document editing} and introduce \textbf{\datasetname}, a corpus of 20K documents with 50K editing instructions and 87K layer-aligned steps, along with the task-specific evaluation protocol \textbf{\evalmetrics} and novelly designed \textbf{\evalscore}.

\item \textbf{Comprehensive Evaluation.} We benchmark 14 open-source and 2 closed-source systems, identifying consistent challenges in instruction following, layout fidelity, and coordination across layers.

\item \textbf{Method and Results.} We propose \textbf{\methodname}, which combines a GRPO-trained multimodal reasoner with a pluggable layer-wise editor. \methodname~ demonstrates strong instruction adherence and layout consistency, surpassing open-source baselines and performing competitively with closed-source systems.
\end{itemize}
\section{Related Work}
\label{sec:related-work}

\subsection{Multi-layer Transparent Image Generation}
Prior research on multi-layer design documents has largely concentrated on the problem of generation. To support this direction, existing datasets are commonly constructed either by extracting layered assets from large-scale image corpora (e.g., LAION~\citep{schuhmann2022laion5b}, COCO~\citep{lin2014mscoco})~\citep{zhang2023text2layer, huang2024layerdiff, tudosiu2024mulan, gu2024adopd}, or by curating poster- and graphic-style designs from online content platforms~\citep{yamaguchi2021canvasvae, pu2025art}. Building on these datasets, a line of work explores models that jointly perform multi-layer generation and understanding with enhanced reasoning abilities~\citep{deng2025bagel, xiao2025mindomni}, as well as synthetic data pipelines for scalable supervision~\citep{burgert2024magick, chen2025prismlayers}. Several approaches further emphasize coordinated multi-layer outputs, where layers are generated with explicit structural or semantic dependencies~\citep{huang2024layerdiff, pu2025art, chen2025prismlayers}.

More recently, researchers have begun to investigate editability by decomposing a flat RGB image into multiple semantically disentangled RGBA layers. For example, Qwen-Image-Layered~\citep{yin2025qwenimagelayered} learns an end-to-end image-to-layer decomposition model and demonstrates post-hoc edits via manual layer-level operations (e.g., resizing or repositioning selected layers) to reduce visual drift. Despite these advances, such approaches primarily target layer discovery or reconstruction, rather than instruction-driven modification of existing design documents.

In contrast, real-world design workflows typically involve non-expert users iteratively editing existing layered documents under high-level instructions, while preserving global structure and layout consistency. This practical requirement remains largely unaddressed by prior work, revealing a clear gap between current research and real-world usage. To bridge this gap, we introduce \textbf{\datasetname~}, the first benchmark that pairs layered design documents with document-level editing instructions and stepwise, layer-aligned edit traces validated through human evaluation. This benchmark reframes the problem from multi-layer generation to faithful and controllable multi-layer editing. 

\subsection{Reasoning-based Image Generation \& Editing}
Driven by recent advances in large language models (LLMs) and training algorithms~\citep{shao2024deepseekmath,yu2025dapo}, reasoning-oriented image generation and editing have achieved remarkable progress \citep{zhang2025reasongen, duan2025got, jiang2025t2i, wu2025reprompt, guo2025can, pan2025unlocking, jin2024reasonpix2pix, zhang2025r}. Current methods may be classified according to the manner in which reasoning is incorporated into the pipeline: (i) \emph{prompt interpretation}, where the system resolves compositional or implicit semantics in user instructions (\eg, temporal or causal cues) prior to editing \citep{chen2025r2i, sun2025t2i, jin2024reasonpix2pix, zhang2025r}; (ii) \emph{prompt extension}, which augments concise instructions with additional structure (\eg, constraints, spatial hints) to enhance output faithfulness \citep{wu2025reprompt, jiang2025t2i, zhang2025reasongen, duan2025got}; and (iii) \emph{generation-time reasoning}, which introduces self-checking or iterative refinement during synthesis to enforce consistency with requirements \citep{guo2025can, pan2025unlocking}. Nevertheless, these approaches are predominantly built on the assumption of a single, flattened canvas and thus lack \emph{layer-aware} reasoning about hierarchical structure, inter-layer dependencies, and document-level constraints (\eg, text fidelity, non-occluding layout). As a result, even when instructions are correctly interpreted, edits often fail to account for relevant layers or disrupt spatial organization. We introduce \textbf{\methodname}, which formalizes \emph{multi-layer document editing} as a reasoning task and ensures consistency via layer selection, layer-wise editing instruction generation, and layer editing.

\section{\datasetname~}
\label{sec:mlpeditbench}

\subsection{Preliminaries}
We define multi-layer document editing as a two-stage process consisting of reasoning and editing. A document $D$ is represented as an ordered set of transparent layers $\mathcal{L}=\{L_i \in \mathbb{R}^{H \times W \times C}\}_{i=1}^n$, rendered by alpha compositing $D=L_1 \oplus \cdots \oplus L_n$. Given a document-level instruction $I_D$, the reasoning stage is performed by a VLM-based reasoner $\mathcal{R}_\phi(D,I_D)\mapsto\hat{\mathcal{I}}=\{\hat{I}_i\}_{i=1}^n$, which predicts layer-specific instructions where each $\hat{I}_i$ either specifies an edit for layer $L_i$ or is a \texttt{no-op} indicating that the layer should remain unchanged. The editing stage is handled by an image-generation editor $\mathcal{E}(\mathcal{L},I_D,\hat{\mathcal{I}})\mapsto D'$, which updates the document by applying $L_i'=\mathcal{E}(L_i,\hat{I}_i)$ if $\hat{I}_i \neq \texttt{no-op}$, and $L_i'=L_i$ otherwise. The final edited document is then reconstructed in the original order as $D'=L_1' \oplus \cdots \oplus L_n'$. A valid solution must satisfy \emph{instruction compliance} (the output follows the semantics, text, and attributes of $I_D$), \emph{structural fidelity} (the global layout and all non-target content remain intact), and \emph{layer awareness} (all and only the layers in $S^\star$ are modified). For diagnostic evaluation, the benchmark provides gold supervision in the form of $S^\star$ and $\mathcal{I}$, enabling measurement of both document-level success (instruction following and fidelity) and decision quality (correctness of layer selection and alignment). Each benchmark instance is therefore specified by five components: the rendered document $D$, its layer decomposition $\mathcal{L}$, the document-level instruction $I_D$, the gold relevant-layer set $S^\star$, and the layer-wise instructions $\mathcal{I}$. 

Since current open- and closed-source\footnote{We verified that GPT-o3 could complete the task in manual trials, but the model was discontinued before our benchmark was finalized, preventing systematic evaluation.} models do not support multi-image (multi-layer) editing interfaces, we design a practical evaluation protocol that treats each method as a \emph{black-box} editor. Specifically, the model only consumes the rendered document $D$ and instruction $I_D$, and produces an edited output $D'$; layer-wise inputs or edits are \emph{not} required. Even under this simplified setting, existing models fail to reliably follow instructions, preserve layout, or render texts (Table~\ref{tab:model_eval}), underscoring the importance and difficulty of the proposed task: no previous work can fully complete it.
Finally, Table~\ref{tab:data_stats} summarizes the dataset statistics. We also show the distribution of layers per document and prompt lengths in Figure~\ref{fig:layer_distribution} and Figure~\ref{fig:prompt_length_distribution} in the Appendix.

\begin{table}
\caption{Statistics of \datasetname~.}
\vspace{-2mm}
\label{tab:data_stats}
\begin{adjustbox}{width=0.8\linewidth,center}
\begin{tabular}{lcc}
\toprule
\textbf{Aspect} & \textbf{Train} & \textbf{Test} \\
\midrule
Number of design documents           &  17.7k  &  1.9k  \\
Avg. \#layers per doc                &  4.45   &  4.44  \\
Avg. \#layers needing edit per doc   &  1.66   &  1.66  \\
Avg. len of doc-level instruction    & 15.56   & 15.53  \\
Avg. len of layer-wise instruction   & 24.50   & 24.48  \\
\bottomrule
\end{tabular}
\end{adjustbox}
\vspace{-2mm}
\end{table}

\subsection{Dataset Construction Pipeline}
\label{dataseet-construction-pipeline}
The dataset contruction pipeline consists of three steps: data collection, document-level instruction generation and layer-wise instruction generation, with human-in-the-loop validation for the last two steps. Alg.~\ref{alg:data_pipeline} in Appendix \ref{app:dataseet-construction-pipeline} illustrates the overall data creation pipeline. We also introduce the details of human-in-the-loop verification steps in App. \ref{app:human-in-the-loop-data-generation}.

\noindent\textbf{Design document collection and layer consolidation.}
We build our corpus from the public Crello dataset \citep{yamaguchi2021canvasvae}, which provides transparent-background, multi-layer \emph{design} documents represented as $(D,\mathcal{L})$, where $D$ is the rendered document and $\mathcal{L}=\{L_i\}_{i=1}^n$ is its layer decomposition. Crello is chosen because (i) our benchmark targets real-world design workflows with non-expert users, so we exclude datasets with synthetically generated layers (\eg, Magick \citep{burgert2024magick}, PrismLayers \citep{chen2025prismlayers}); and (ii) our focus is on scenarios where text, decorative elements, and imagery interact, so we omit multi-layer resources derived from \emph{natural} images (\eg, MuLAn \citep{tudosiu2024mulan}, MLCID \citep{huang2024layerdiff}). Although ART \citep{pu2025art} introduces a large-scale design corpus, it is not publicly available and thus excluded. To make $\mathcal{L}$ tractable, we apply a \emph{structure-preserving consolidation} procedure $\mathcal{C}(\mathcal{L}) \mapsto \mathcal{L}'$: an MLLM (InternVL3-38B \citep{zhu2025internvl3}) classifies layers into \emph{text}, \emph{decoration}, or \emph{image}, and non-overlapping layers within each category are merged using layout metadata while preserving $z$-order and alpha boundaries. This reduces $|\mathcal{L}|$ (2–50) to a semantically coherent $\mathcal{L}'$ ($|\mathcal{L}'|$ varies 1-12) without discarding content.

\noindent\textbf{Document-level instruction generation.} 
Given a consolidated design document $(D,\mathcal{L})$, we generate a document-level instruction $I_D$ for each item. We adopt a two-stream pipeline that balances diversity and realism. (i) \textit{Persona-based stream:} six personas $p_j \sim \text{PersonaHub}$ are sampled, and InternVL3-38B generates candidate instructions ${I_D^{(j)}}$ by adapting $D$ to each persona’s domain while preserving its design intent (\eg, “concert poster’’ $\rightarrow$ “historical exhibition poster’’, $p_j$ is a “historian’’). (ii) \textit{Document-based stream:} the model proposes semantically proximal domain transfers grounded in $D$ itself (\eg, “summer camp’’ $\rightarrow$ “winter camp’’). The combined candidate pool ${I_D^{(j)}}$ is then ranked by clarity, specificity, and realism, with low-quality cases removed through lightweight automatic filtering and regeneration until criteria are met. Finally, a human-in-the-loop validation stage ensures applicability and removes instructions that are infeasible, yielding the final $I_D$.

\noindent\textbf{Layer-wise instruction generation.}
For each benchmark instance $(D,I_D,\mathcal{L})$, we provide a set of \emph{layer-aligned} editing instructions $\mathcal{I}={I_i}$ specifying how each relevant layer should be modified to realize the document-level intent. During document-level instruction synthesis, the InternVL3-38B is simultaneously prompted to produce step-wise edits as a program that decomposes $I_D$ into atomic actions (\eg, “replace text "piano concert" with "historical exhibition"”). We then align steps to layers using a novel MLLM-based content-aware matcher to produce layer-wise instructions ${I_i}$. The matching algorithm is detailed in Appendix~\ref{app:steps-layer-matcher-algorithm}. Finally, automatically generated instructions are filtered by rule-based validators and refined through human-in-the-loop expert review, ensuring clarity, feasibility, and faithfulness to real design workflows. The resulting edited layers $S^\star$ and aligned instructions $\mathcal{I}$ thus combine automated alignment with human refinement to provide reliable gold supervision.
\section{Benchmarking with \datasetname~}

\subsection{\textbf{\evalmetrics~}}
\label{sec:evaluation-metrics}
For a comprehensive assessment of our benchmark, we introduce \textbf{\evalmetrics~}, which encompasses four key evaluation dimensions: instruction following, layout consistency, aesthetics, and text rendering. To holistically reflect model performance on the task, we further integrate the four perceptual criteria into a unified score, denoted as \textbf{\evalscore~}.

\noindent\textbf{Instruction Following.}  
To assess whether the model faithfully executes an editing instruction $I_D$, we design a VQA-style evaluation metric. Given the document $D$, the target layer $S^\star$, and its layer-specific prompt $\mathcal{I}$, InternVL3-38B is prompted to generate a question–answer pair for each edited layer. Each question explicitly grounds the edit in spatial, textual, or entity-level detail (\eg, \textit{``Has the main image be changed to a museum scene?''}), with a binary answer of ``yes'' or ``no.'' The instruction-following score is defined as the proportion of edits judged correct across all layers.

\noindent\textbf{Layout Consistency.}
To evaluate structural fidelity, we measure layout consistency between original and edited documents using mask-level representations. We extract spatial masks $\mathcal{M} = \{M_i\}$ and $\mathcal{M}' = \{M'_j\}$ using Adopd Doc2Mask model \citep{gu2024adopd} from the original document $D$ and edited document $D'$, then we design a new matching algorithm to match the two sets of spatial masks. The detailed calculation function is shown in Appendix \ref{appendix:layout-consistency-evaluation}.

\noindent\textbf{Aesthetics.}
We assess whether edits preserve or improve overall visual appeal using a frozen aesthetics predictor (\emph{Aesthetic Predictor V2.5} \citep{aesthetic_v2_5}). We directly utilize the score as final evaluation.

\noindent\textbf{Text Rendering.}  
We evaluate the \emph{faithfulness} of edited text with an OCR–VQA pipeline. Specifically, we first apply the Adopd Doc2BBox model \citep{gu2024adopd} to detect text regions in the edited image $L_j'$, and then use InternVL3-38B to extract the corresponding text $t'$. Given the instruction $I_D$, we prompt the MLLM to verify whether $t'$ satisfies the required edit, producing a score in $\{0, 0.5, 1\}$. Unlike conventional text-alignment metrics (\eg, SentenceBERT~\citep{Reimers2019SentenceBERTSE}), our approach does not assume a unique ground truth: multiple valid edits may satisfy $I_D$, and thus a judgment-based evaluation better captures instruction faithfulness.

\noindent\textbf{\evalscore~.} Although the four evaluation dimensions comprehensively capture different aspects of the multi-layer design document editing task, they cannot be treated as independent objectives. 
For example, if an editing model fails to modify the document and simply outputs the unedited input, the \textit{layout consistency} score would reach 100\%, while \textit{instruction following} and \textit{text rendering} would be zero. 
In this case, the high layout consistency is meaningless, since it does not indicate a successful edit. 
To jointly model the interdependence among these factors, we introduce \textbf{\evalscore}, a unified metric that aggregates the four perceptual criteria into a single holistic score.
Let the raw scores of \textit{instruction following (IF)}, \textit{layout consistency (LC)}, \textit{text rendering (TR)}, and \textit{aesthetics (A)} be normalized to $[0,1]$ as:
\begin{equation}
\text{IF}_h = \frac{\text{IF}}{100}, \quad
\text{LC}_h = \frac{\text{LC}}{100}, \quad
\text{TR}_h = \frac{\text{TR}}{100}, \quad
\text{A}_h = \frac{\text{A}}{10}.
\end{equation}

We employ an instruction-following–based \textbf{sigmoid gate} to control the influence of other metrics:
\begin{equation}
g(\text{IF}_h) =
\frac{\sigma(k(\text{IF}_h - \tau)) - \sigma(-k\tau)}
{\sigma(k(1-\tau)) - \sigma(-k\tau)}, 
\quad 
\sigma(x) = \frac{1}{1 + e^{-x}},
\end{equation}
where $\tau$ defines the gate threshold and $k$ controls the steepness. 
A higher $\tau$ makes the gate stricter, while a larger $k$ sharpens the transition. The overall \textbf{\evalscore} is computed as:
\begin{equation}
\begin{aligned}
\text{MiLDEScore} =
&~ w_{\text{if}} \text{IF}_h + w_{\text{tr}} \text{TR}_h \\
&+ g(\text{IF}_h)\left(w_{\text{lc}}\text{LC}_h 
+ w_{\text{a}}\text{A}_h\right) \\
&+ w_{\text{sy}}\, g(\text{IF}_h)\, \text{IF}_h\, \text{LC}_h.
\end{aligned}
\end{equation}

The sigmoid gate $g(\text{IF}_h)$ ensures that \textit{layout consistency} and \textit{aesthetics} only contribute meaningfully when the instruction-following score is sufficiently high. 
When the model fails to follow the editing instruction ($\text{IF}_h < \tau$), the gate value remains near zero, effectively suppressing irrelevant high LC or A scores. 
As $\text{IF}_h$ increases, these terms are gradually activated, allowing models that both follow instructions and preserve layout to achieve higher overall scores. The last term $w_{\text{sy}} g(\text{IF}_h)\text{IF}_h\text{LC}_h$ captures the synergy between instruction accuracy and spatial consistency. 
It provides an additional reward when both metrics are simultaneously high, reflecting the natural coupling between semantic correctness and visual coherence in human judgment. More details are discussed in App. \ref{app:details-in-evalscore}.

\noindent\textbf{Layer Decision Accuracy.} In addition to metrics for edited document quality, we also incorporate another metric called layer decision accuracy. As shown in Figure \ref{fig:multi-layer-editing}, in many cases in our benchmark, not all layers require modification. Therefore, we additionally report the layer decision accuracy to measure whether the model can correctly identify which layers should be edited. 

\subsection{Evaluation and Analysis}

To conduct evaluation on \textbf{\datasetname~}, we conduct comprehensive evaluation on 14 open-source models, with 12 reasoning-free models and 2 reasoning-enhanced models, and 2 closed-source models. Note that in these experiments, we only take design document $D$ and document-level editing instruction $I_D$ as input, because current models cannot conduct multiple layer editing sinmutaneously. Specifically, the task here is $\mathcal{E}(D,I_D) \rightarrow D'$. The primary results are presented in Table \ref{tab:model_eval}. 
Please refer to Appendix \ref{appendix:experiments} for detailed experiment and evaluation setup.

\begin{table*}[t!]
\scriptsize
\centering
\caption{Evaluation results of different models. Instruction Fol., Layout Cons., Text Rend., and Layer Dec. Acc. represents information following, layout consistency, text rendering and layer decision accuracy, respectively. 
For all scores, higher values indicate better performance. The highest score for \colorbox{backred!50}{closed-source} and \colorbox{backblue!75}{open-source} text-to-image models are marked in red and blue respectively, and \underline{underline} represents the second in open-source models. Note that for previous baselines incapable of multi-layer editing, the \textit{layer decision accuracy} metric is not applicable.
}
\label{tab:model_eval}
\begin{tabular}{c|l|*{6}{>{\centering\arraybackslash}p{1.7cm}}}
\toprule
\textbf{\texttt{\#}} & \textbf{Model} & 
\shortstack{Instruction Fol.} & 
\shortstack{Layout Cons.} & 
\shortstack{Aesthetic} & 
\shortstack{Text Rend.} &
\shortstack{\textbf{\evalscore~}} & 
\shortstack{Layer Dec. Acc.} \\
\midrule
\rowcolor{gray!8}\multicolumn{8}{l}{\textit{\textbf{Open-source Models}}} \\ 
\midrule
\texttt{1} & Instruct-Pix2Pix~\citep{brooks2023instructpix2pix} & 2.30 & \colorbox{backblue!75}{93.46} & \underline{4.23} & 17.16 & 6.23 & -- \\
\texttt{2} & MagicBrush~\citep{zhang2023magicbrush} & 7.37 & 72.08 & 3.68 & 16.60 & 8.47 & -- \\
\texttt{3} & UniWorld-v1~\citep{lin2025uniworld} & 5.75 & 61.59 & 3.91 & 22.04 & 9.15 & -- \\
\texttt{4} & ICEdit~\citep{zhang2025context} & 2.28 & 64.60 & 3.42 & 18.25 & 6.43 & -- \\
\texttt{5} & UltraEdit~\citep{zhao2024ultraedit} & 12.41 & 85.31 & 3.54 & 11.39 & 10.35 & -- \\
\texttt{6} & AnyEdit~\citep{yu2025anyedit} & 6.51 & 56.73 & 3.96 & 21.83 & 9.40 & -- \\
\texttt{7} & OmniGen~\citep{xiao2025omnigen} & 3.83 & 85.96 & 3.90 & 19.76 & 7.73 & -- \\
\texttt{8} & Qwen-Image-Edit~\citep{wu2025qwen} & 10.09 & 74.20 & 4.12 & 24.32 & 12.42 & -- \\
\texttt{9} & Flux1-Kontext~\citep{batifol2025flux} & 12.49 & 48.32 & 3.94 & 19.31 & 11.58 & -- \\
\texttt{10} & VAREdit~\citep{mao2025visual} & 6.60 & 68.10 & 3.18 & 9.49 & 5.86 & -- \\
\texttt{11} & Step1X-Edit~\citep{liu2025step1x} & 6.56 & 84.09 & 3.98 & 18.70 & 8.84 & -- \\
\texttt{12} & Bagel~\citep{deng2025bagel} & \underline{14.23} & 48.59 & 3.54 & 13.49 & 10.80 & -- \\
\midrule
\rowcolor{gray!8}\multicolumn{8}{l}{\textit{\textbf{Reasoning-enhanced Models}}} \\ 
\midrule
\texttt{13} & Step1X-Edit w/ Thinking & 10.48 & 82.16 & 4.11 & \underline{28.67} & 14.17 & -- \\
\texttt{14} & Bagel w/ Thinking & 13.60 & 60.91 & 3.65 & 14.51 & 11.23 & -- \\
\midrule
\rowcolor{gray!8}\multicolumn{8}{l}{\textit{\textbf{Closed-source Models}}} \\ 
\midrule
\texttt{15} & GPT-Image-1~\citep{openai_gpt_image1} & \colorbox{backred!50}{25.46} & 36.24 & \colorbox{backred!50}{4.66} & 39.67 & 25.60 & -- \\
\texttt{16} & Nano Banana~\citep{gemini-nanobanana} & 24.04 & \colorbox{backred!50}{58.42} & 4.52 & \colorbox{backred!50}{40.32} & \colorbox{backred!50}{27.10} & -- \\
\midrule
\rowcolor{gray!8}\multicolumn{8}{l}{\textit{\textbf{\methodname~} (Ours)}} \\
\midrule
\texttt{18} & Qwen2.5VL-3B + Flux & 13.29 & 90.15 & \colorbox{backblue!75}{4.32} & 27.52 & \underline{16.10} & 42.90 \\
\texttt{19} & Qwen2.5VL-7B + Flux & \colorbox{backblue!75}{20.71} & \underline{93.24} & 4.19 & \colorbox{backblue!75}{36.75} & \colorbox{backblue!75}{25.90} & \colorbox{backblue!75}{80.46} \\
\bottomrule
\end{tabular}
\end{table*}

\noindent\textbf{Finding 1: Current image editing models struggle with design document editing.}  
Our evaluation reveals that both open-source and closed-source models exhibit certain limitations in instruction following and text rendering. For open-source models (\texttt{\#1}-\texttt{\#14}), the average instruction-following accuracy is only about 10\%, meaning that in nearly 90\% of cases the specified edits are not correctly executed. Even the strongest closed-source baseline, GPT-Image-1 (\texttt{\#15}), achieves only 25.46\% instruction following accuracy, underscoring the substantial gap between current image editing capabilities and the demands of multi-layer document editing in realistic scenarios.

\noindent\textbf{Finding 2: Closed-source models achieve stronger instruction following but sacrifice format consistency.}  
Closed-source models substantially outperform open-source ones in instruction following, text rendering, and aesthetics. 
For example, in terms of instruction-following accuracy for content editing, GPT-Image-1 (\texttt{\#15}) surpasses the best-performing open-source model Bagel (\texttt{\#12}) by 78\% (25.46\% vs. 14.23\%).
For text-rendering score in content editing, Nano Banana (\texttt{\#16}) exceeds the best-performing open-source model Step1X-Edit w/ Thinking (\texttt{\#13}) by 40.6\% (40.32\% vs. 28.67\%).
However, these gains come at the expense of layout-consistency.
In particular, GPT-Image-1 (\texttt{\#15}) achieves the lowest score in layout-consistency,
and Nano Banana (\texttt{\#16}) performs only on par with the open-source average. 
Notably, the comparably high layout-consistency scores in open-source models often stem from trivial artifacts, such as outputting the unedited document, which preserves layout without satisfying the instruction. This highlights a critical trade-off: closed-source models follow instructions more reliably, but they lack the ability to maintain structural fidelity in design documents, which is a limitation with significant consequences for real-world editing workflows.

\vspace{-1mm}
\noindent\textbf{Finding 3: Reasoning-enhanced models provide only marginal gains for document editing.}  
Augmenting open-source editors with explicit reasoning mechanisms (``w/ Thinking'') yields limited improvements. Step1X-Edit w/ Thinking (\texttt{\#13} vs. \texttt{\#11}) improves instruction-following accuracy from 6.56\% to 10.48\% 
and achieves the highest text-rendering score (28.67\%), suggesting that reasoning can help decompose instructions into more precise edits. However, Bagel w/ Thinking (\texttt{\#14} vs. \texttt{\#12}) decreases instruction-following accuracy from 14.23\% to 13.60\% and provides no substantial gains in other metrics. Overall, the benefits remain modest relative to the difficulty of the task. Current reasoning modules primarily capture textual intent but struggle to ground edits within multi-layer document structures, especially when document-level editing prompts usually represents editing text and image sinmutaneously. This underscores the need for deeper multimodal reasoning integration, rather than shallow textual planning, to advance design document editing.

\begin{figure*}[t!]
\centering
\includegraphics[width=\textwidth]{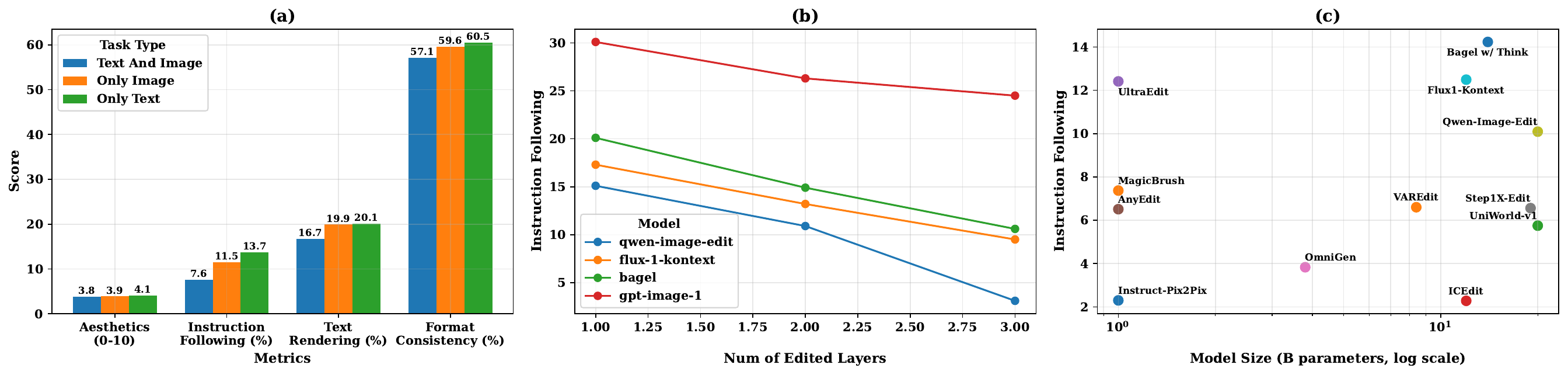}
\caption{(a) Evaluation metrics with editing type. (b) Instruction following score with number of edited layers. (c) Instruction following score with model size.
}
\label{fig:finding4} 
\end{figure*}

\noindent\textbf{Finding 4: Complex reasoning paths exacerbate editing errors.} 
\noindent\textbf{Finding 4: Complex reasoning paths exacerbate editing errors.} Model performance degrades markedly as editing complexity increases. First, we sampled 150 cases from test set and classify them into three types based on the editing domain: text-only, image-only, and text+image editing. We report the average content editing score of three open-source models (Qwen-Image-edit, Flux1-Kontext, and Bagel). As shown in Figure~\ref{fig:finding4} (a), instruction-following drops from 13.7\% (text-only) and 11.5\% (image-only) to 7.6\% (text+image), with parallel declines in text rendering, aesthetics, and format consistency. Figure~\ref{fig:finding4} (b) further reveals a strong effect of layer depth: Bagel falls from 20.1\% (one layer) to 10.6\% (three layers), Flux1-Kontext from 17.3\% to 9.5\%, and Qwen-Image-Edit from 15.1\% to 3.1\%; even GPT-Image-1 drops from 30.1\% to 24.5\%. Finally, Figure~\ref{fig:finding4} (c) shows that larger model size does not consistently improve performance. In summary, performance degrades as editing complexity increases—both across modalities and with deeper layer structures—highlighting that current models struggle to reason over complex editing intents. Moreover, scaling model size does not consistently yield improvements, suggesting that advancing multimodal \emph{reasoning capability} is crucial for progress in design document editing.

\vspace{-2mm}
\section{The \methodname~ Framework}

Recognizing the reasoning inaccuracies, layout consistency issue and the fundamental problem that current image editing model cannot do multiple layer editng, we propose \textbf{\methodname~}, consisting of an RL-trained reasoner and a frozen image editor. Specifically, our agent receives a design document $D$ with multiple transparent background layers $\mathcal{L}$ and a document-level instruction $I_D$, and then produce $D'$ by editing \emph{exactly} the relevant layers and re-compositing them in the original $z$-order. Specifically, the task here is $\textit{Agent}(D,I_D,\mathcal{L}) \rightarrow (D', \mathcal{L}')$. We introduce our agent in Section \ref{sec:mildeagent} and evaluate on our benchmark on Section \ref{sec:agent-evaluation-results}. We also discuss human evaluation in Appendix \ref{app:human-evaluation}.

\subsection{Reasoning-Guided Multi-Layer Document Editing}\label{sec:mildeagent}

\begin{figure*}[t!]
\centering
\includegraphics[width=0.95\textwidth]{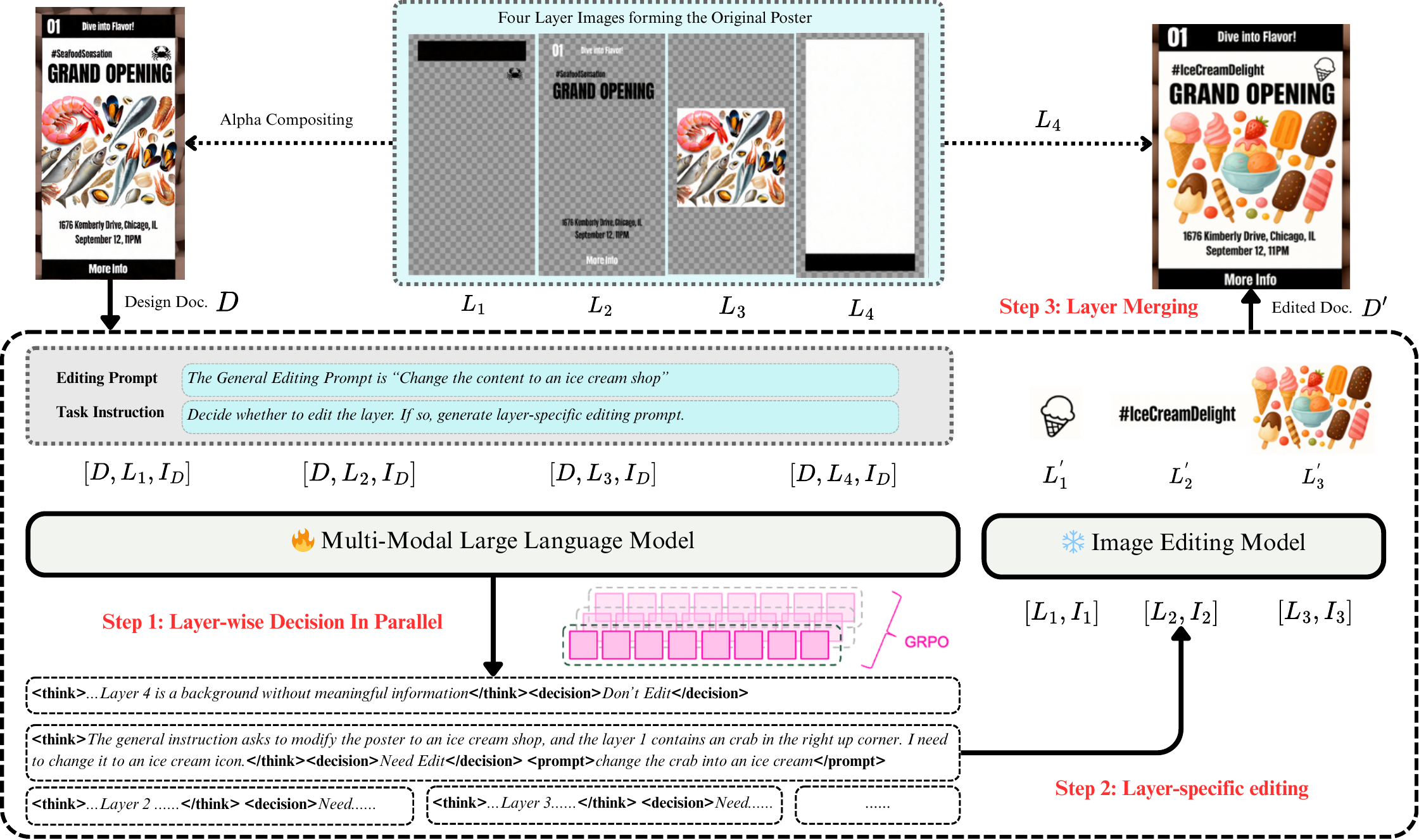}
\caption{The illustration of \methodname~.}
\label{fig:mlpeagent}
\end{figure*}

Our \textbf{\methodname} is a two-stage framework for multi-layer document editing, where the reasoner $\mathcal{R}_\phi$ performs instruction decomposition and the editor $\mathcal{E}$ performs layer-wise editing. The details of our agent is illustrated in Figure \ref{fig:mlpeagent}.

\textbf{Reasoning.} The reasoning stage is handled by a VLM-based reasoner $\mathcal{R}_\phi$, which takes $(D,L_i,I_D)$ as input and outputs for each layer a binary decision $y_i \in \{0,1\}$ and, if $y_i=1$, a \emph{layer-conditioned prompt} $I_i$. To train $\mathcal{R}_\phi$, we adopt Group Relative Policy Optimization (GRPO) \citep{shao2024deepseekmath}, a RL method that evaluates groups of sampled responses, computes relative advantages by normalizing their rewards, and applies a clipped KL-regularized objective. This design reduces variance in credit assignment and encourages the model to distinguish between relatively better and worse responses, which is particularly beneficial for structured reasoning tasks (see Appendix \ref{sec:preliminary} for details).

Following this paradigm, we design a task-specific per-layer reward to supervise $\mathcal{R}_\phi$. The outputs of the reasoner must follow a structured format:
\begin{equation}
\begin{aligned}
\texttt{<think>} \ldots \texttt{</think>}
\texttt{<decision>} \ldots \texttt{</decision>} \\
\texttt{<prompt>} \ldots \texttt{</prompt>}
\end{aligned}
\end{equation}
\hspace{-0.2em}where the three segments denote hidden reasoning, the binary decision $y_i$, and the layer-conditioned prompt $I_i$, respectively. The per-layer reward $\mathcal{R}_i$ then consists of three components:
\vspace{-5mm}
\begin{equation}
\begin{aligned}
r_f &= \mathbb{1}[\text{format is valid}], \quad
r_d = \mathbb{1}[y_i = y_i^\star], \\
r_p &= \text{BLEU}(I_i, I_i^\star) \in [0,1].
\end{aligned}
\end{equation}
The final per-layer reward is defined as
\begin{align}
\mathcal{R}_i =
\begin{cases}
(r_f + r_d + r_p)/3, & r_d=1, \\[4pt]
(r_f + r_d)/2, & r_d=0~.
\end{cases}
\end{align}
where $r_f$ verifies syntactic correctness, $r_d$ measures decision accuracy against the gold label 
$y_i^\star=\mathbb{1}[L_i \in S^\star]$, and $r_p$ evaluates prompt quality relative to the reference 
instruction $I_i^\star$. The prompt reward $r_p$ is only applied when the decision is correct ($r_d=1$).

\textbf{Editing.} The editing stage uses a frozen image-generation editor $\mathcal{E}$ for stability and modularity. For each selected layer $L_i$ ($y_i=1$), a binary mask $M_i$ is extracted from its alpha channel (optionally refined with region cues), and the editor updates it as $L_i'=\mathcal{E}(L_i,I_i,M_i)$. For non-selected layers ($y_i=0$), no operation is applied and $L_i'=L_i$. Transparency is preserved by restoring the original alpha to unedited regions. The final document is reconstructed by alpha compositing $D'=L_1'\oplus L_2'\oplus\cdots\oplus L_n'$, where $\oplus$ denotes standard alpha blending, ensuring global layout consistency while fulfilling the document-level instruction $I_D$.

\subsection{Experimental Results}
\label{sec:agent-evaluation-results}

\noindent\textbf{Setup.} We incorporate one of the SOTA MLLM, \texttt{QwenVL2.5-3B/7B} \citep{bai2025qwen2} as our reasoner, and applied GRPO algorithm to train on content editing tasks, with a freezed \texttt{Flux-1-Kontext} as editing model. The rollout number is set to 5 and the batch size to 512. All experiments are conducted on 8 A100 GPUs. 

\begin{figure*}[t!]
\centering
\includegraphics[width=0.95\textwidth]{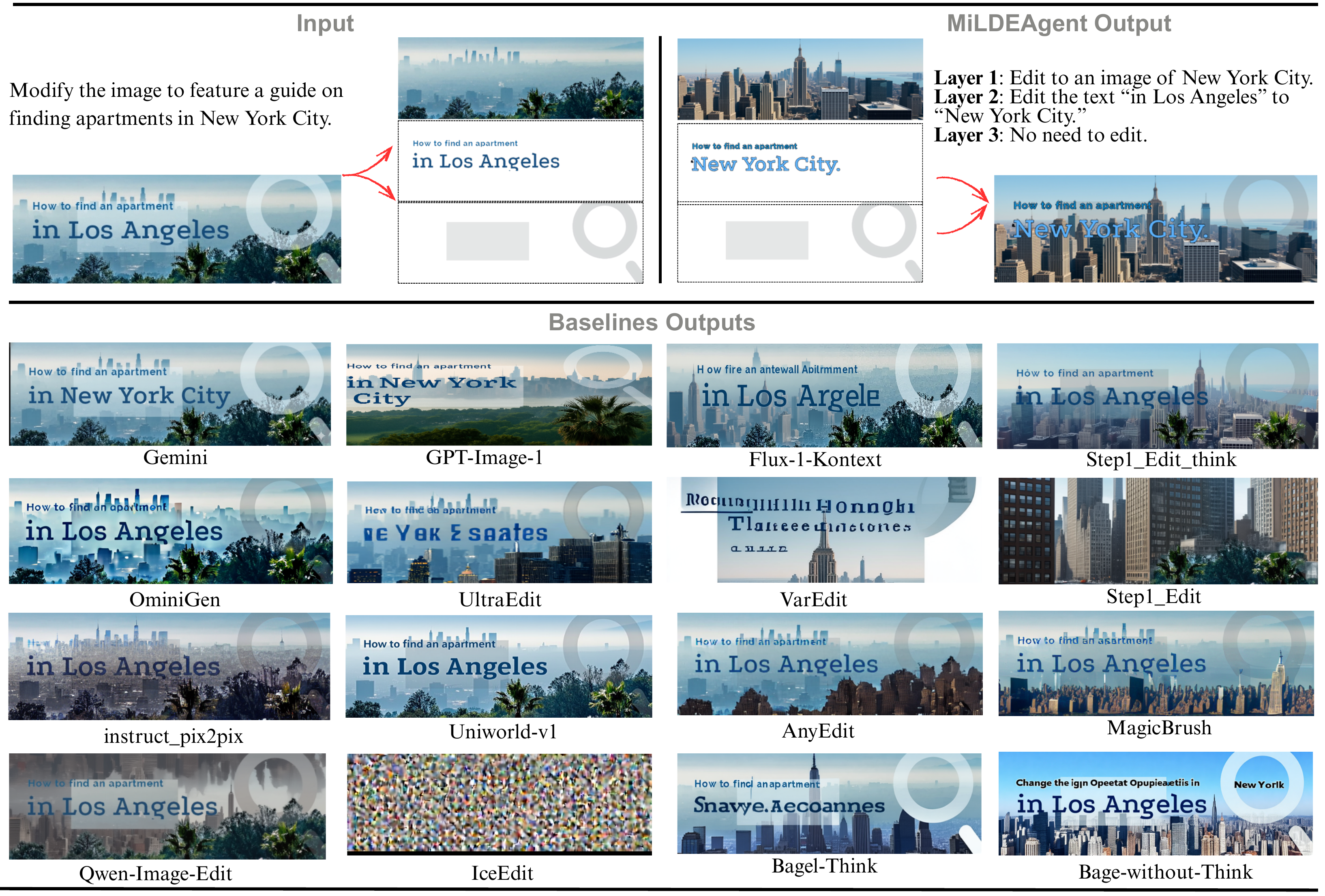}
\caption{Qualitative comparison results between \methodname~ and other baselines.}
\label{fig:qualitative_analysis}
\vspace{-5mm}  
\end{figure*}

\noindent\textbf{Quantitative Results.} As shown in Table ~\ref{tab:model_eval}, our proposed \methodname~ significantly outperforms all baselines in the content editing regime. Specifically, \methodname~ achieves 25.9\% in \evalscore, representing a 82.78\% improvement over the strongest open-source baseline (Bagel, 14.17\%) and narrowing the gap with closed-source systems (Nano Banana, 27.1\%) and even outperforming GPT-Image-1 (25.6\%) while preserving editability. Independently, on instruction following, 7B version \methodname~ outperforms all open-source baselines. On format consistency, \methodname~ maintains 93.2\%, rivaling the best-performing diffusion-based editors and exceeding closed-source models by over +30 points. Importantly, our agent exhibits strong text rendering performance (36.8\%), surpassing all open-source baselines ($\leq$ 24.3\%) and approaching commercial systems (40\%). This highlights the effectiveness of our reasoning-based approach in handling multi-layer textual elements, a persistent weakness of prior methods. Finally, on layer decision accuracy (80.5\%), \methodname~ demonstrates robust layer-aware reasoning, an ability entirely absent from existing baselines, thereby validating the necessity of reasoning-enhanced frameworks for this task. 
Taken together, these results establish that multi-layer document editing requires explicit reasoning mechanisms, rather than relying solely on generation or editing heuristics.  \methodname~ consistently balances instruction fidelity, fine-grained textual rendering, and layer-aware decomposition, making it the first system to robustly address multi-layer editing at scale.

\noindent\textbf{Quality Analysis.} As illustrated in Figure~\ref{fig:qualitative_analysis}, the input design document consists of three layers. 
Our agent successfully identifies that the first layer, which contains the background image of Los Angeles, 
should be edited to depict New York City. 
In addition, the text ``in Los Angeles'' in the second layer is correctly modified to ``New York City.'' 
The third layer, however, is purely decorative and is correctly recognized as not requiring any modification. 
After applying an open-source image editing model, our agent composites the edited layers 
with the unedited ones to form the final output. 
The resulting image preserves the original layout while accurately updating the relevant content, 
and it also retains per-layer information for future flexible modifications by users. 
In contrast, all other baselines fail in this task, even under single-image editing settings. 
For instance, Gemini only changes the textual content without modifying the background image, 
whereas GPT-Image-1 fails to maintain layout consistency. 
Other open-source baselines either fail to edit the text (e.g., OmniGen, Step1-Edit) 
or completely fail to perform meaningful edits (e.g., VarEdit, IceEdit).

\noindent\textbf{Failure Cases.} 
Our agent still exhibits certain failure modes. 
First, as layer decisions are made independently, multiple layers may occasionally be edited simultaneously, 
resulting in unintended overlaps or visual conflicts. 
Second, even with high layer decision accuracy (e.g., the 7B model), 
the overall instruction-following score can remain low due to 
(i) ambiguous or underspecified layer-wise editing prompts, and 
(ii) the inherent limitations of the underlying image editing model. 
A potential solution is to integrate a self-checking mechanism that verifies the merged output 
and re-invokes editing when inconsistencies are detected. 
Further analyses and examples are provided in Appendix~\ref{app:failure-cases}.

\section{Conclusion}
In this work, we introduced \datasetname~, the first benchmark for reasoning-based multi-layer poster editing, together with a novel evaluation metrics. Through comprehensive experiments, we demonstrated that existing methods struggle to accurately edit posters based on general simple editing prompt. To address these limitations, we proposed \methodname~, which leverages a GRPO-trained reasoner for layer selection and prompt generation, coupled with a open-source image editor, significantly improving reasoning ability and editing quality.  
{
    \small
    \bibliographystyle{ieeenat_fullname}
    \bibliography{main}
}
\clearpage
\setcounter{page}{1}
\maketitlesupplementary

\section{MiLDEBench}

\subsection{Data Generation Pipeline}
\label{app:dataseet-construction-pipeline}

\begin{algorithm}
  \scriptsize
  \caption{\small{Data Construction Pipeline}}
  \label{alg:data_pipeline}
  \SetKwInOut{Input}{Input}
  \SetKwInOut{Output}{Output}
  \Input{Design document $D$ with layers $\mathcal{L}$}
  \Output{Validated document-level instruction $I_D$, layer-wise instructions $\mathcal{I}=\{I_i\}$, edited layers $S^\star$}

  \BlankLine
\textbf{Part A: Document-level Instruction Generation}
\begin{steplist}
  \item Generate candidate instructions $\{I_D^j\}$ from $D$ via personas $p_j \sim \text{PersonaHub}$;
  \item Rank and filter $\{I_D^j\}$ by clarity, realism, and consistency;
  \item Human validation $\Rightarrow$ finalize $I_D$.
\end{steplist}

\BlankLine
\textbf{Part B: Layer-wise Instruction Generation}
\begin{steplist}
  \item Decompose $I_D$ into step-wise edits $\mathcal{A}=\{a_j\}$;
  \item Match each $a_j$ to candidate layers $L_k \in \mathcal{L}$ using content-aware alignment;
  \item Form preliminary instructions $I_k$ and filter by clarity, feasibility, and consistency;
  \item Human validation $\Rightarrow$ finalize $\mathcal{I}$ and relevant-layer set $S^\star$.
\end{steplist}

  \end{algorithm}

\begin{figure*}[t]
    \centering
    \begin{subfigure}[t]{0.48\linewidth}
        \centering
        \includegraphics[width=\linewidth]{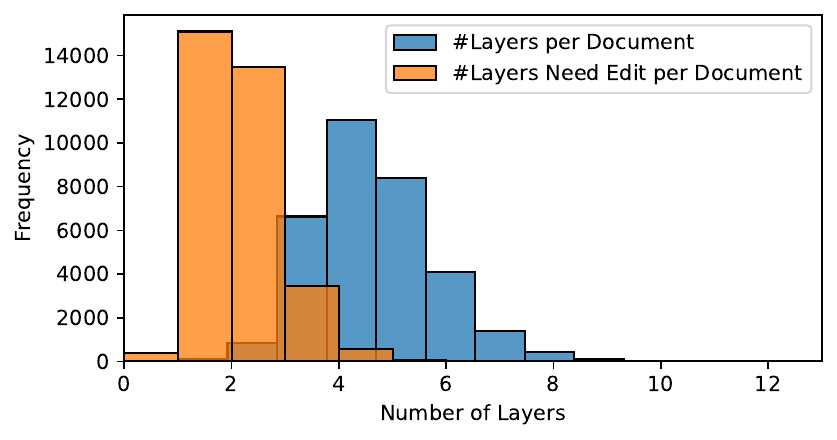}
        \subcaption{Train set \#layer distribution.}
        \label{fig:image1}
    \end{subfigure}
    \hfill
    \begin{subfigure}[t]{0.48\linewidth}
        \centering
        \includegraphics[width=\linewidth]{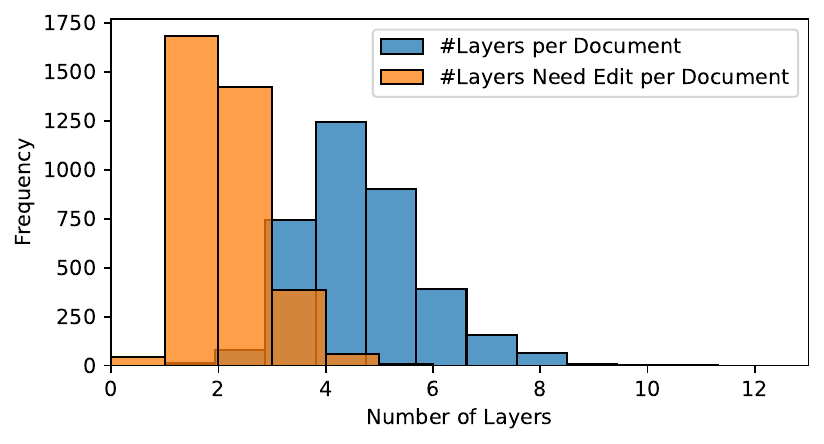}
        \subcaption{Test set \#layer distribution.}
        \label{fig:image2}
    \end{subfigure}
    \caption{Distributions of the total number of layers per document and the number of layers requiring edits per document in the \datasetname~.}
    \label{fig:layer_distribution}
\end{figure*}

\begin{figure*}[t]
    \centering
    \begin{subfigure}[t]{0.48\linewidth}
        \centering
        \includegraphics[width=\linewidth]{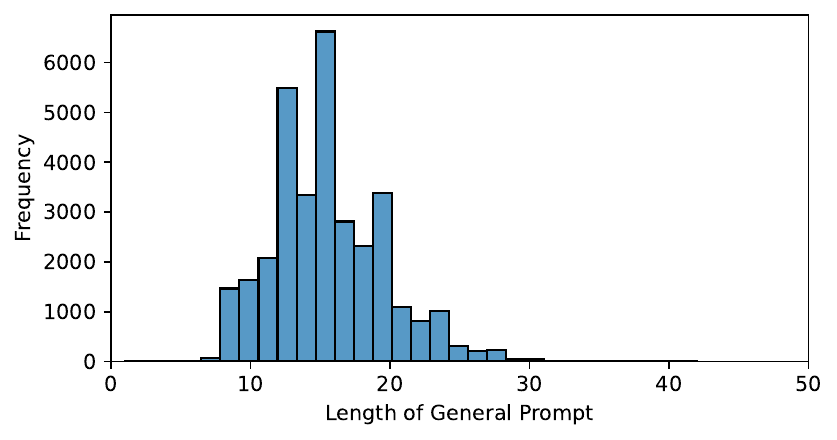}
        \subcaption{Distribution of general prompt length.}
        \label{fig:image1}
    \end{subfigure}
    \hfill
    \begin{subfigure}[t]{0.48\linewidth}
        \centering
        \includegraphics[width=\linewidth]{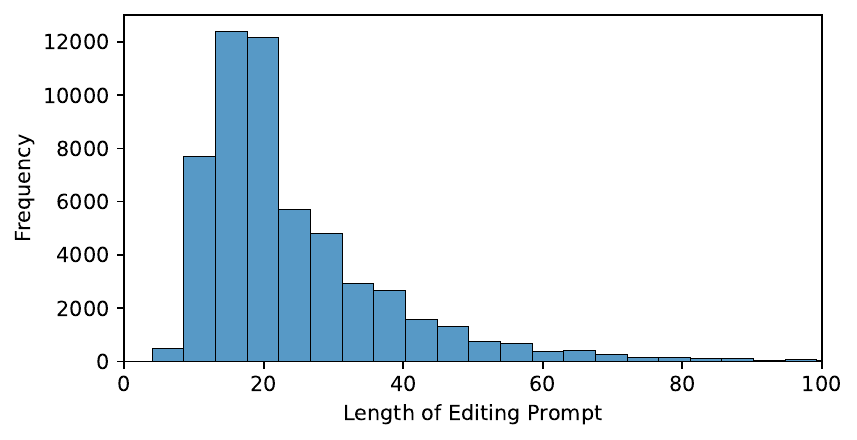}
        \subcaption{Distribution of editing prompt length.}
        \label{fig:image2}
    \end{subfigure}
    \caption{Distributions of general prompt lengths and the editing prompt lengths in the \datasetname~.}
    \label{fig:prompt_length_distribution}
\end{figure*}

\subsection{Layer-wise Instruction Generation}
\label{app:steps-layer-matcher-algorithm}
In this section, we describe the matcher used to align step-wise editing prompts with document layers. Given a set of step-wise prompts ${\mathcal{I}_k}$ and the layer set ${S_j}$ with known types (textual or visual), we first classify each prompt $\mathcal{I}_k$ using InternVL3-38B into either a \emph{text-editing} or an \emph{image-editing} category. A prompt is considered eligible only for layers of the corresponding type (i.e., text prompts for textual layers, image prompts for visual layers). Within each category, we process prompts sequentially: for each $\mathcal{I}_k$, we traverse the candidate layers in $z$-order and query InternVL3-38B to assess whether $\mathcal{I}_k$ semantically applies to $S_j$. Upon a positive match, $\mathcal{I}_k$ is assigned to $S_j$, and the procedure advances to the next prompt. This iterative matching continues until all prompts have been assigned or no valid layer remains.

\subsection{Human-in-the-loop Quality Control}
\label{app:human-in-the-loop-data-generation}
For each generated editing instruction, we ask human annotators to check whether the editing instruction is reasonable based on the design document content. If not, we filter out this example.

\section{\evalmetrics~}

\subsection{Layout Consistency}
\label{appendix:layout-consistency-evaluation}
To evaluate structural fidelity, we measure layout consistency between original and edited documents using mask-level representations. We extract spatial masks $\mathcal{M} = \{M_i\}$ and $\mathcal{M}' = \{M'_j\}$ using Adopd Doc2Mask model \citep{gu2024adopd} from the original document $D$ and edited document $D'$, then we design a new matching algorithm to match the two sets of spatial masks.
For matched pairs, we assess position consistency (normalized centroid displacement), shape consistency (IoU), and area consistency (size ratio). Unmatched layers incur area-proportional penalties, with deleted layers penalized more heavily than newly created ones. The final score combines matching rate, average consistency scores, and penalty deductions with empirically tuned weights, providing a comprehensive measure of layout preservation robust to structural variations.

To assess the \textbf{structural fidelity} requirement---specifically whether the edited document $D'$ preserves the spatial arrangement and geometric relationships of elements---we introduce a comprehensive layout consistency metric that operates on mask-level representations of document layers. Given the inherent challenges of multi-layer editing where the number of layers may change ($|\mathcal{L}'| \neq |\mathcal{L}|$) and layer correspondences may be disrupted due to editing operations, our evaluation framework employs a principled matching strategy followed by multi-dimensional consistency assessment.

\textbf{Mask Extraction and Matching.} For both the original document $D$ and edited document $D'$, we extract layer-wise masks $\mathcal{M} = \{M_i\}_{i=1}^{|\mathcal{L}|}$ and $\mathcal{M}' = \{M'_j\}_{j=1}^{|\mathcal{L}'|}$ respectively using Adopd Doc2Mask model \cite{gu2024adopd}, where each mask $M_i \in [0,1]^{H \times W}$ represents the spatial footprint of layer $L_i$. To establish correspondences between original and edited layers, we formulate mask matching as a bipartite graph optimization problem: we compute a pairwise IoU similarity matrix $\mathbf{S} \in \mathbb{R}^{|\mathcal{L}| \times |\mathcal{L}'|}$ where $S_{ij} = \text{IoU}(M_i, M'_j)$, then apply the Hungarian algorithm to find the optimal matching $\mathcal{P}^* = \arg\max_{\mathcal{P}} \sum_{(i,j) \in \mathcal{P}} S_{ij}$ subject to IoU threshold filtering ($S_{ij} \geq \tau_{\text{IoU}}$).

\textbf{Multi-Dimensional Consistency Assessment.} For each matched pair $(M_i, M'_j) \in \mathcal{P}^*$, we evaluate three complementary aspects of layout preservation: (1) \textbf{Position consistency} measures centroid displacement normalized by image diagonal: $c_{\text{pos}}(M_i, M'_j) = 1 - \frac{\|\text{centroid}(M_i) - \text{centroid}(M'_j)\|_2}{\sqrt{H^2 + W^2}}$; (2) \textbf{Shape consistency} directly uses the IoU between masks: $c_{\text{shape}}(M_i, M'_j) = \text{IoU}(M_i, M'_j)$; (3) \textbf{Area consistency} computes the ratio of smaller to larger mask areas: $c_{\text{area}}(M_i, M'_j) = \frac{\min(\text{area}(M_i), \text{area}(M'_j))}{\max(\text{area}(M_i), \text{area}(M'_j))}$.

\textbf{Unmatched Layer Penalty.} To account for layers that appear or disappear during editing, we introduce a penalty mechanism that distinguishes between disappeared layers (present in $\mathcal{L}$ but unmatched in $\mathcal{L}'$) and newly created layers (present in $\mathcal{L}'$ but unmatched in $\mathcal{L}$). The penalty for each unmatched layer is proportional to its normalized area, with disappeared layers receiving full penalty and new layers receiving a reduced penalty (coefficient 0.7) to reflect that layer creation may be intentional: $p_{\text{disappeared}} = \sum_{i \in \mathcal{U}_{\text{orig}}} \text{area}(M_i)$ and $p_{\text{new}} = 0.7 \sum_{j \in \mathcal{U}_{\text{edit}}} \text{area}(M'_j)$, where $\mathcal{U}_{\text{orig}}$ and $\mathcal{U}_{\text{edit}}$ denote unmatched layer indices.

\textbf{Final Score Computation.} The overall layout consistency score aggregates matched-layer performance with unmatched-layer penalties:
\begin{align}
\text{LayoutConsistency} &= \max\bigg(0, \omega_{\text{match}} \cdot r_{\text{match}} \\
&+ \omega_{\text{pos}} \cdot \bar{c}_{\text{pos}} + \omega_{\text{shape}} \cdot \bar{c}_{\text{shape}} \nonumber \\
&+ \omega_{\text{area}} \cdot \bar{c}_{\text{area}} \\
&- \omega_{\text{penalty}} \cdot (p_{\text{disappeared}} + p_{\text{new}})\bigg),
\end{align}
where $r_{\text{match}} = \frac{|\mathcal{P}^*|}{\max(|\mathcal{L}|, |\mathcal{L}'|)}$ is the matching rate, $\bar{c}_{\cdot}$ denotes average consistency scores across matched pairs, and $\{\omega_{\cdot}\}$ are empirically set weights (0.25, 0.2, 0.2, 0.2, 0.15 respectively). This metric provides a comprehensive assessment of layout preservation that is robust to layer count variations and sensitive to both geometric distortions and structural changes.

\section{Experiments}
\label{appendix:experiments}

\subsection{Baseline Models}
\noindent\textbf{Baseline Open-source Models} We evaluate on 14 open-source models covering auto regressive and diffusion-based framework. The model size ranges from 1B to 20B. The details of each model are shown in Table \ref{tab:model_stats}.

\begin{table}[t!]
\small
\centering
\caption{Introduce of each baseline model. Rea.-En. represents whether the model is reasoning-enhanced.
}
\label{tab:model_stats}
\begin{tabular}{lccccc}
\toprule
Model & Size & Type & Rea.-En. \\ 
\midrule
Instruct-Pix2Pix \cite{brooks2023instructpix2pix} & 1B & Diffusion & \xmark \\
MagicBrush \cite{zhang2023magicbrush} & 1B & Diffusion & \xmark \\
UniWorld-v1 \cite{lin2025uniworld} & 20B & Diffusion & \xmark \\
ICEdit \cite{zhang2025context} & 12B & Diffusion & \xmark \\
UltraEdit \cite{zhao2024ultraedit} & 1B & Diffusion & \xmark \\
AnyEdit \cite{yu2025anyedit} & 1B & Diffusion & \xmark  \\
OmniGen \cite{xiao2025omnigen} & 3.8B & Diffusion & \xmark \\
Step1X-Edit \cite{liu2025step1x} & 19B & Diffusion & \xmark \\
Qwen-Image-Edit \cite{wu2025qwen}& 20B & Diffusion & \xmark \\
Flux1-Kontext \cite{batifol2025flux} & 12B & Diffusion & \xmark \\
Bagel w/o Think \cite{deng2025bagel} & 14B & Diffusion & \xmark \\
Bagel w/ Think \cite{deng2025bagel} & 14B & Diffusion & \cmark \\
VAREdit \cite{mao2025visual} & 8.4B & AR & \xmark \\
DIM-Edit \cite{zeng2025draw} & 4.6B & Diffusion & \xmark \\
\bottomrule
\end{tabular}
\end{table}

\subsection{\evalscore~}
In Table \ref{tab:model_eval}, we set $\tau=0.3$, $k=10.0$, $w_{if} = 0.30$, $w_{lc} = 0.30$, $w_{tr} = 0.30$, $w_a = 0.10$, and $w_{sy} = 0.15$. We discuss the reason we chose these in Section \ref{app:details-in-evalscore} and \ref{app:ablation-study-on-weight-parameters}.

\section{\methodname~}

\subsection{Preliminary of GRPO Algorithm}
\label{sec:preliminary}
Group Relative Policy Optimization (GRPO) \citep{shao2024deepseekmath} has been proved to be helpful for improving reasoning capabilities for LLM \citep{shao2024deepseekmath}, Multi-modal understanding \citep{huang2025vision} and even image generation \citep{zhang2025reasongen, jiang2025t2i}. GRPO computes advantages from a group of responses. Given each question-anwer pair $(q, a)$, old policy $\pi_{\theta_{\text{old}}}$ randomly samples $G$ responses, denoted as $\{o_i\}_{i=1}^G$. Each response $o_i$ is then fed into a reward model to obtain a reward $R_i$. Then, the advantage of the $i$-th response is obtained by normalizing the rewards of the group:

\begin{equation}
A_i = \frac{\mathcal{R} - \text{mean}(\{\mathcal{R}_i\}_{i=1}^G)}{\text{std}(\{\mathcal{R}_i\}_{i=1}^G)}
\end{equation}

GRPO applies a clipped objective similar to PPO with a KL penalty term:

\begin{equation}
\begin{aligned}
\mathcal{J}_{\mathrm{GRPO}}&(\theta) = \mathbb{E}_{(q, a) \sim \mathcal{D}, \{o_i\}_{i=1}^G \sim \pi_{\theta_{\text{old}}}(\cdot \mid q)} \Bigg[
\frac{1}{\sum_{i=1}^G |o_i|} \\
&\sum_{i=1}^G \sum_{t=1}^{|o_i|}
\Big( \min \big( r_{i,t}(\theta)\hat{A}_i,
\operatorname{clip}(r_{i,t}(\theta), 1-\varepsilon, 1+\varepsilon)\hat{A}_i \big)
\\
& - \beta D_{\mathrm{KL}}(\pi_\theta \,\|\, \pi_{\mathrm{ref}}) \Big)
\Bigg]
\end{aligned}
\end{equation}
where $r_{i,t}(\theta)$ is the important weight for each token $t$:
\begin{equation}
r_{i, j}(\theta)=\frac{\pi_\theta\left(o_{i, j} \mid q, o_{i,<j}\right)}{\pi_{\theta_{\text {old }}}\left(o_{i, t} \mid q, o_{i,<j}\right)} .
\end{equation}

Usually in the reasoning task with only textual output, the model is asked to generate responses following a structured format. The total rewards consists of two rule-based rewards: (1) format reward and the accuracy of the specific downstream task.

\subsection{Ablation Study}
\begin{figure}[t!]
\centering
\includegraphics[width=0.8\linewidth]{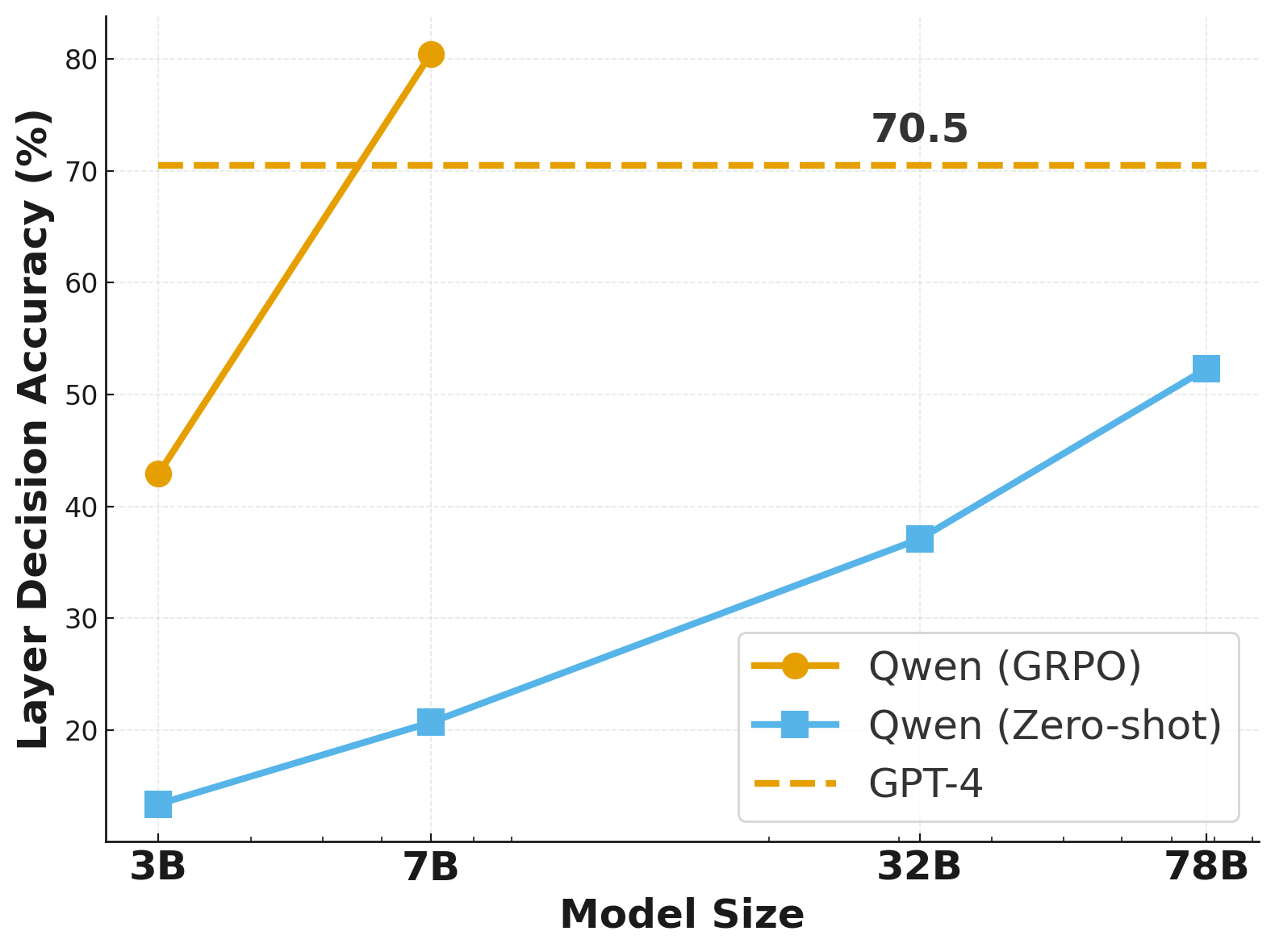}
\caption{Layer decision accuracy with model size.}
\label{fig:agent-ablation}
\vspace{-5mm}  
\end{figure}

\noindent\textbf{Ablation 1: GRPO-trained reasoner outperforms all zero-shot models in layer decision accuracy.} Reasoner is the key of \methodname~, therefore, we conduct ablation study on the RL-trained reasoner with other larger open-/closed-source MLLMs on layer decision accuracy metrics.
As shown in Figure \ref{fig:agent-ablation} (b), we observe that models equipped with a GRPO-trained reasoner consistently surpass their zero-shot counterparts across all tested scales. For instance, QwenVL2.5-7B with GRPO achieves 80.5\% accuracy, compared to only 20.7\% for its zero-shot variant, a nearly 4$\times$ improvement. Similarly, QwenVL2.5-3B with GRPO improves from 13.4\% to 42.9\%, highlighting that structured reinforcement-style reasoning is beneficial even at smaller scales. Strikingly, the 7B GRPO-trained model not only outperforms all zero-shot baselines—including much larger 32B and 78B models—but also slightly outperforms GPT-4. These results underscore that \emph{reasoning-oriented training, rather than model scaling alone, is the dominant factor for reliable layer decision making}, establishing GRPO as a crucial ingredient for advancing multi-layer document editing.


\noindent\textbf{Ablation 2: Image editing model also influence the final performance.} In this experiment, we randomly select 100 samples from content editing test set and utilize the GRPO-trained QwenVL2.5-7B model as reasoner to test different image editing models. As shown in Table~\ref{tab:ablation-image-editing}, although all models achieve broadly comparable scores, systematic differences emerge across evaluation dimensions. GPT-Image-1 consistently achieves the best overall results, with 19.7\% in instruction following, 4.4\% in aesthetics, and 30.6\% in text rendering, outperforming the best open-source alternatives by a clear margin. Among open-source models, Qwen-Image-Edit exhibits relatively stronger instruction following and text rendering, while Bagel and Flux1-Kontext are more balanced but weaker in fidelity and reasoning. These results indicate that even with the same reasoning mechanism, the fidelity and controllability of the editing backbone strongly shape the final quality of document editing. Consequently, improvements in low-level editing architectures are complementary to reasoning-based approaches, and both are required to achieve robust performance in multi-layer editing. One thing to mention is that in Table \ref{tab:model_eval}, we choose Flux-1-Kontext as our main image editing model in general. We acknowledge that using more powerful image editing model will improve the performance, but this will not influence tha main conclusion of our findings.

\begin{table}[h]
\centering
\caption{Evaluation score with different image editing models. The reasoner model is GRPO-trained Qwen2.5-VL-7B. The workflow is same as \methodname~.}
\label{tab:ablation-image-editing}
\begin{tabular}{lccc}
\toprule
Model & IF & Aes. & TR \\
\midrule
Qwen-Image-Edit & 18.5 & 4.0 & 26.4 \\
Flux-1-Kontext & 17.9 & 3.65 & 23.5 \\
Bagel & 17.3 & 3.76 & 25.4 \\
GPT-Image-1 & 19.7 & 4.37 & 30.6 \\
\bottomrule
\end{tabular}
\end{table}

\subsection{Human Evaluation}
\label{app:human-evaluation}
\noindent\textbf{Setup.} We sampled 100 instances from \textbf{\datasetname~} as a subset for human evaluation. To save evaluation time, we select Nano Banana, Bagel w/ Thinking, Flux1-Kontext and UltraEdit as baselines for human evaluation which covers closed-source models, reasoning-enhanced models and open-source models. For our agent, we choose Qwen2.5VL-7B model to evaluate. In this way, we have 500 data points where each baseline has inference results on 100 instances. For each data point, we have two different annotators who are Ph.D. or master's students or with expertise in multimodal domains, or professional designer knowing much on document designing to give ratings independently. We adopt the same evaluation criteria as \datasetname~, where for each sample, the annotators need to give a score for the four aspects: instruction following, layout consistency, aesthetic, and text rendering. Besides, we also ask each annotators to evaluate on the overall quality considering all aspects together as the overall assessment. We use a scale of $\{0, 1,2,3\}$ for each aspect to saving annotation time, where 1, 2, and 3 indicate the quality is bad, fair, and good, respectively.

\begin{table*}[t!]
\centering
\caption{Human evaluation results of existing baseline models on \textbf{\datasetname~}. IF, LC, Aes. TR, OQ represents instruction following, layout consistency, aesthetic, text rendering, and overall quality respectively.}
\label{table:human-eval}
\begin{tabular}{c|l|*{5}{>{\centering\arraybackslash}p{1.7cm}}}
\toprule
\textbf{\texttt{\#}} & \textbf{Model} & 
\shortstack{IF} & 
\shortstack{LC} & 
\shortstack{Aes.} & 
\shortstack{TR} &
\shortstack{OQ} \\
\midrule
\rowcolor{gray!8}\multicolumn{7}{l}{\textit{\textbf{Baselines}}} \\ 
\midrule
\texttt{1} & Instruct-Pix2Pix~\citep{brooks2023instructpix2pix} & 0.05 & 2.85 & 1.23 & 0.67 & 0.27 \\
\texttt{2} & MagicBrush~\citep{zhang2023magicbrush} & 0.35 & 2.27 & 1.07 & 0.59 & 0.28 \\ 
\midrule
\texttt{14} & Bagel w/ Thinking & 0.87 & 2.05 & 1.16 & 0.58 & 0.39 \\
\midrule
\texttt{16} & Nano Banana~\citep{gemini-nanobanana} & 1.34 & 1.76 & 1.95 & 1.84 & 1.45 \\
\midrule
\rowcolor{gray!8}\multicolumn{7}{l}{\textit{\textbf{\methodname~} (Ours)}} \\
\midrule
\texttt{19} & Qwen2.5VL-7B + Flux & 1.28 & 2.83 & 1.27 & 1.75 & 1.37 \\
\bottomrule
\end{tabular}
\end{table*}

\noindent\textbf{Results.} We show the human evaluation results in Table \ref{table:human-eval}. The human evaluation is generally consistent with the automatic evaluation in Table \ref{tab:model_eval}. Our \methodname performs comparable with closed-source model Nano Banana and significantly outperforms than open-source models. We report the Inter Annotator Agreement (IAA) in Table \ref{tab:iaa}. The inter-annotator agreement is good. Because the aesthetic is subjective and open-ended, the agreement score is relatively lower than other scores.

\begin{table*}[t!]
\caption{\textbf{Inter-Annotator Agreement} of human evaluation in terms of Cohen's Kappa score. Please note that overall quality is corresponding to \evalscore~.}
\label{tab:iaa}
\centering
\begin{tabular}{c c c c c c c}
\toprule
\textbf{Instruction Following} & \textbf{Layout Consistency} & \textbf{Aesthetic} & \textbf{Text Rendering} & \textbf{Overall Quality} & \textbf{AVG} \\
\midrule
0.75 & 0.71 & 0.61 & 0.72 & 0.69 & 0.70 \\
\bottomrule
\end{tabular}
\end{table*}

\subsection{More Details in \evalscore~}
\label{app:details-in-evalscore}
\noindent\textbf{Motivation and Rescaling.} The motivation to design the \evalscore~ is to find a comprehensive metrics considering all aspects in design document editing that can overall assess the quality of the edited document. Another reason is that we should not consider each criteria separately. For example, one model may have very high layout consistency score but low instruction following score. This means that the model fails to edit the document or directly return the original document to users. In this way, high layout consistency score is meaningless. In order to aggregate the four aspects together, we need to first scale them into the same scope. According to the Aesthetic model \citep{aesthetic_v2_5}, the scope is ranging from 1 to 10, while other three aspects ranging from 1 to 100. Therefore, we rescale them into the same scope by dividing aesthetic score by 10 and dividing other three scores by 100. 


\noindent\textbf{Other Baselines.} 
There exist multiple ways to aggregate the four metrics into an overall score. 
We compare our proposed method with four representative baselines: 
(1) \textbf{DW\_sum (Direct Weighted Sum)}, 
(2) \textbf{GeoMean (Geometric Mean Aggregation)}, and 
(3) \textbf{HCoreSup (Harmonic Core–Support Aggregation)}. 
Each baseline captures different assumptions about metric interactions.

\noindent\textbf{(1) Direct Weighted Sum (DW\_sum).} 
The most straightforward way is a linear weighted combination of the normalized scores:
\[
S_{\text{DW}} = w_{if}\cdot IF_h + w_{tr}\cdot TR_h + w_{lc}\cdot LC_h + w_{a}\cdot A_h.
\]
This method assumes each metric contributes independently and linearly. 
Although simple and smooth, it tends to overestimate models that exhibit high layout consistency but poor instruction following, failing to penalize unedited outputs.

\noindent\textbf{(2) Geometric Mean (GeoMean).} 
The geometric mean combines all criteria multiplicatively:
\[
S_{\text{GM}} = \Big((IF_h)^{w_{if}}\cdot (TR_h)^{w_{tr}}\cdot (LC_h)^{w_{lc}}\cdot (A_h)^{w_a}\Big)^{1/\sum w},
\]
which enforces that any low-dimensional score (e.g., a very low $IF_h$) will significantly lower the final score. 
This method penalizes unbalanced models but may underestimate systems that excel in one dimension while being average in others, leading to overly conservative evaluation.

\noindent\textbf{(3) Harmonic Core–Support (HCoreSup).} 
We divide metrics into ``core'' (\emph{instruction following, text rendering}) and ``support'' (\emph{layout consistency, aesthetics}) groups:
\[
\begin{aligned}
S_{\text{HC}} = 
\frac{2 \cdot S_{\text{core}} \cdot S_{\text{sup}}}{S_{\text{core}} + S_{\text{sup}}}, 
\quad
S_{\text{core}} &= \frac{w_{if}\cdot IF_h + w_{tr}\cdot TR_h}{w_{if}+w_{tr}},\\
S_{\text{sup}} &= \frac{w_{lc}\cdot LC_h + w_{a}\cdot A_h}{w_{lc}+w_{a}}.
\end{aligned}
\]
This harmonic mean encourages balanced performance between content correctness and visual consistency, while still allowing partial compensation between the two groups.

As is shown in this table, our sigmoid-gated synergistic method achieves the highest consistency with human ratings, showing that incorporating soft gating and interaction terms better captures subjective quality assessment.

\subsection{Ablation Study on Weight Parameters}
\label{app:ablation-study-on-weight-parameters}
To validate the effectiveness of our proposed evaluation metric and determine the optimal weight configuration, we conduct comprehensive ablation studies on the weight parameters. Our evaluation score is formulated as:
\begin{equation}
\begin{aligned}
\text{Score} &= w_{if} \cdot \hat{IF} + w_{tr} \cdot \hat{TR} \\&+ g \cdot (w_{lc} \cdot \hat{LC} + w_a \cdot \hat{A}) + w_{sy} \cdot g \cdot \hat{IF} \cdot \hat{LC}
\end{aligned}
\end{equation}
where $\hat{IF}$, $\hat{LC}$, $\hat{TR}$, and $\hat{A}$ denote the normalized scores for Instruction Following, Local Consistency, Text Rendering, and Aesthetics, respectively. The gating function $g = \sigma(k(\hat{IF} - \tau))$ modulates the contribution of consistency and aesthetics based on instruction following performance, with $\tau = 0.3$ and $k = 10.0$. The synergy term $w_{sy} \cdot g \cdot \hat{IF} \cdot \hat{LC}$ captures the multiplicative interaction between instruction following and local consistency.

\textbf{Experimental Setup.} We systematically evaluate different weight configurations while satisfying the constraint $w_{if} + w_{lc} + w_{tr} + w_a = 1$. To assess the alignment between our automatic metric and human judgment, we compute the Spearman rank correlation coefficient ($\rho$) between the scores produced by each configuration and human evaluation scores collected from expert annotators.

\begin{table*}[t]
\centering
\caption{Ablation study on weight parameters and comparison of different scoring functions. All configurations satisfy $w_{if} + w_{lc} + w_{tr} + w_a = 1$. For MiLDEScore, the synergy weight $w_{sy}$ is fixed at 0.15 unless otherwise noted. $\rho$ denotes Spearman correlation with human evaluation. DW\_sum, GeoMean, and HCoreSup do not utilize the synergy term.}
\label{tab:ablation}
\resizebox{\textwidth}{!}{
\begin{tabular}{lccccc|cccc}
\toprule
\multirow{2}{*}{\textbf{Configuration}} & \multirow{2}{*}{$w_{if}$} & \multirow{2}{*}{$w_{lc}$} & \multirow{2}{*}{$w_{tr}$} & \multirow{2}{*}{$w_a$} & \multirow{2}{*}{$w_{sy}$} & \multicolumn{4}{c}{\textbf{Spearman $\rho$}} \\
\cmidrule(lr){7-10}
& & & & & & MiLDEScore & DW\_sum & GeoMean & HCoreSup \\
\midrule
\textbf{Ours (Optimal)} & \textbf{0.30} & \textbf{0.30} & \textbf{0.30} & \textbf{0.10} & \textbf{0.15} & \textbf{0.88} & 0.58 & 0.61 & 0.79 \\
\midrule
\multicolumn{10}{l}{\textit{Varying Primary Weights}} \\
IF Dominant (High) & 0.45 & 0.25 & 0.20 & 0.10 & 0.15 & 0.82 & 0.61 & \textbf{0.64} & 0.80 \\
IF Dominant (Mid) & 0.40 & 0.25 & 0.25 & 0.10 & 0.15 & 0.85 & \textbf{0.62} & 0.63 & \textbf{0.81} \\
LC Dominant (High) & 0.25 & 0.45 & 0.20 & 0.10 & 0.15 & 0.81 & 0.43 & 0.48 & 0.70 \\
LC Dominant (Mid) & 0.25 & 0.40 & 0.25 & 0.10 & 0.15 & 0.84 & 0.47 & 0.51 & 0.72 \\
TR Dominant (High) & 0.20 & 0.25 & 0.45 & 0.10 & 0.15 & 0.79 & 0.49 & 0.53 & 0.71 \\
TR Dominant (Mid) & 0.25 & 0.25 & 0.40 & 0.10 & 0.15 & 0.83 & 0.52 & 0.55 & 0.74 \\
A Dominant & 0.25 & 0.25 & 0.25 & 0.25 & 0.15 & 0.76 & 0.41 & 0.46 & 0.65 \\
Equal Weights & 0.25 & 0.25 & 0.25 & 0.25 & -- & 0.76 & 0.45 & 0.52 & 0.71 \\
\midrule
\multicolumn{10}{l}{\textit{Varying Synergy Weight (MiLDEScore only)}} \\
No Synergy & 0.30 & 0.30 & 0.30 & 0.10 & 0.00 & 0.813 & -- & -- & -- \\
Low Synergy & 0.30 & 0.30 & 0.30 & 0.10 & 0.05 & 0.842 & -- & -- & -- \\
High Synergy & 0.30 & 0.30 & 0.30 & 0.10 & 0.25 & 0.856 & -- & -- & -- \\
Very High Synergy & 0.30 & 0.30 & 0.30 & 0.10 & 0.30 & 0.831 & -- & -- & -- \\
\bottomrule
\end{tabular}
}
\end{table*}
\textbf{Results and Analysis.} As shown in Table~\ref{tab:ablation}, our optimal configuration ($w_{if} = 0.30$, $w_{lc} = 0.30$, $w_{tr} = 0.30$, $w_a = 0.10$, $w_{sy} = 0.15$) achieves the highest Spearman correlation of $\rho = 0.908$ with human evaluation, significantly outperforming alternative configurations. We analyze the impact of each design choice:

\textbf{(1) Balanced vs. Dominant Weights.} Configurations that heavily favor a single dimension (IF Dominant, LC Dominant, or TR Dominant with weights of 0.45) yield substantially lower correlations ($\rho = 0.650$), indicating that no single metric alone captures the multifaceted nature of image editing quality. Similarly, the Equal Weights configuration ($w_{if} = w_{lc} = w_{tr} = w_a = 0.25$) achieves only $\rho = 0.671$, suggesting that treating aesthetics equally with other dimensions does not align well with human preferences.

\textbf{(2) Role of the Synergy Term.} The synergy term proves crucial for capturing the interaction between instruction following and local consistency. Removing this term entirely (No Synergy, $w_{sy} = 0$) reduces the correlation to $\rho = 0.692$, while excessive synergy weighting (High Synergy, $w_{sy} = 0.30$) yields a similar degradation ($\rho = 0.692$). This demonstrates that moderate synergy ($w_{sy} = 0.15$) effectively models how human evaluators reward edits that simultaneously follow instructions accurately and maintain visual coherence.

\textbf{(3) Comparison with Alternative Scoring Functions.} 
We further compare \evalscore~ against three baseline aggregation methods: Direct Weighted Sum (DW\_sum), Geometric Mean (GeoMean), and Harmonic Core-Support (HCoreSup). As shown in Table~\ref{tab:ablation}, \evalscore~ consistently outperforms all baselines across different weight configurations. Under the optimal setting, \evalscore~ achieves $\rho = 0.88$, substantially surpassing HCoreSup ($\rho = 0.79$), GeoMean ($\rho = 0.61$), and DW\_sum ($\rho = 0.58$). 

The performance gap stems from two key innovations in \evalscore~: (i) the \textit{adaptive gating mechanism} that dynamically modulates the contribution of visual quality metrics based on instruction following performance, preventing inflated scores for models that preserve content without executing the requested edit; and (ii) the \textit{synergy term} that explicitly captures the positive interaction between instruction following and local consistency, which none of the baselines can model. Notably, even the best-performing baseline configurations (IF Dominant Mid for DW\_sum and HCoreSup, IF Dominant High for GeoMean) achieve at most $\rho = 0.81$, still significantly below our optimal \evalscore~. This consistent advantage across all configurations demonstrates that the architectural design of \evalscore~, rather than parameter tuning alone, accounts for its superior alignment with human judgment.

\subsection{Failure Cases}
\label{app:failure-cases}
One example is shown in Figure \ref{fig:more-examples-1}. Our agent successfully predicts whether the layer should be edited. However, the merged document shows overlapped text and main image. This can be partially solved by self-checking mechanism in future. However, adding self-checking mechanism is not the main story of our paper, therefore, we leave this part as our future plan.

\section{More Cases}
\begin{figure*}[t!]
\centering
\includegraphics[width=\textwidth]{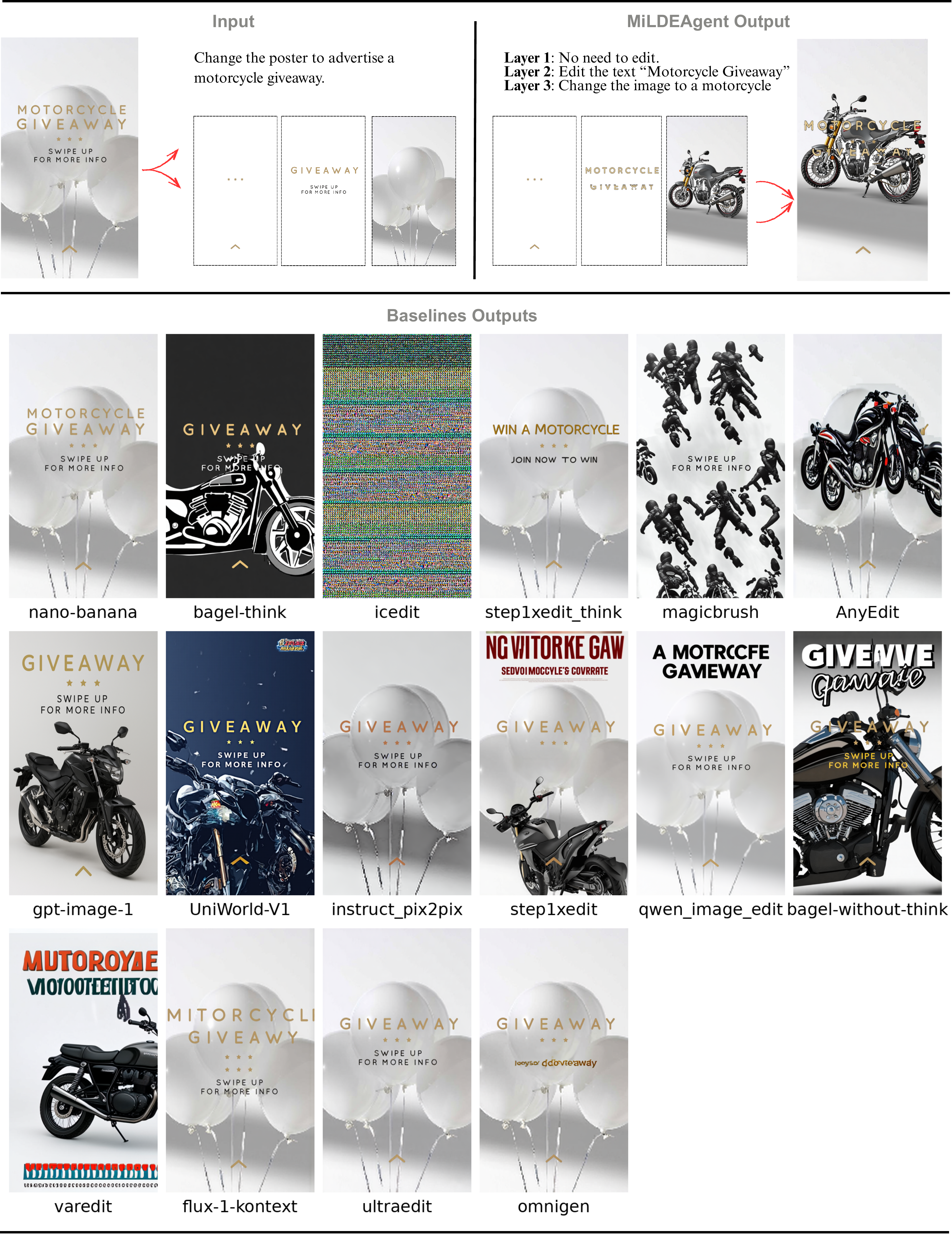}
\caption{More examples 1.
}
\label{fig:more-examples-1} 
\end{figure*}

\begin{figure*}[t!]
\centering
\includegraphics[width=\textwidth]{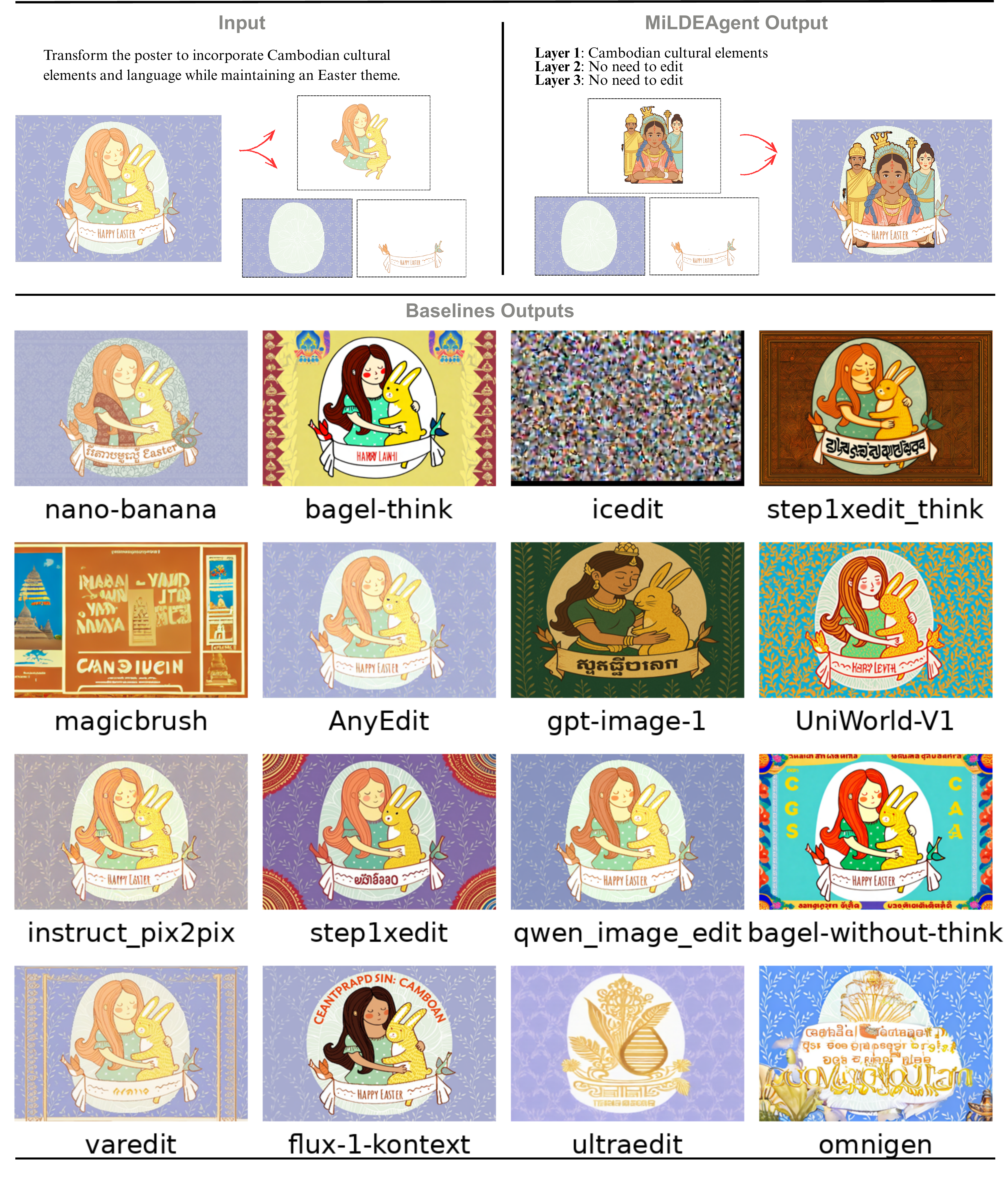}
\caption{More examples 2.
}
\label{fig:more-examples-2} 
\end{figure*}

\begin{figure*}[t!]
\centering
\includegraphics[width=\textwidth]{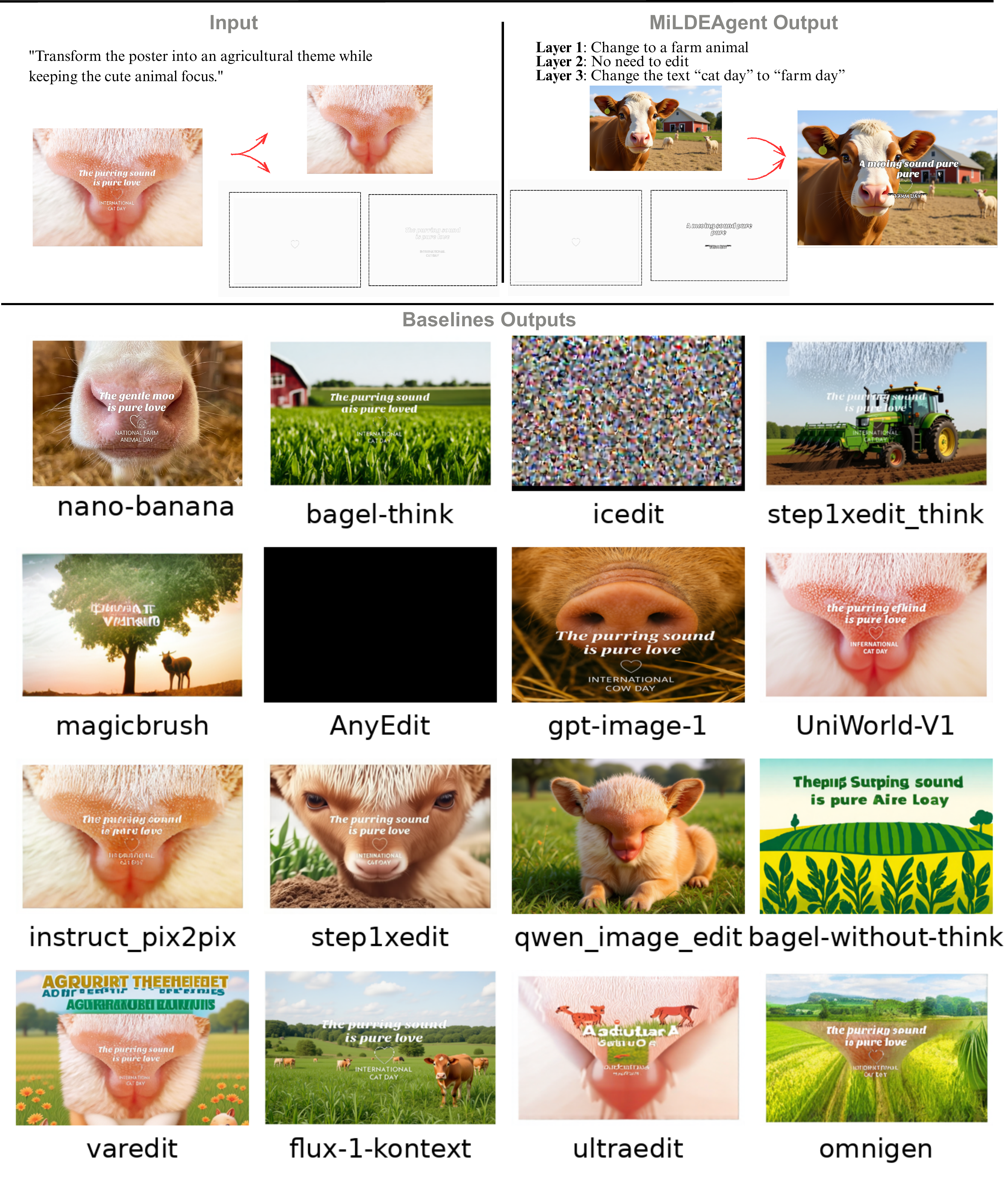}
\caption{More examples 3.
}
\label{fig:more-examples-3} 
\end{figure*}

We show more cases from Figure \ref{fig:more-examples-1} to \ref{fig:more-examples-3}.

\end{document}